\RequirePackage{fix-cm}
\documentclass[twocolumn]{svjour3}          
\smartqed  
\usepackage{graphicx}
\usepackage{latexsym}
\usepackage{url}
\usepackage{hyperref}
\usepackage{amsmath}
\usepackage{multirow}
\usepackage{amssymb}
\usepackage{siunitx}
\usepackage{cite}
\usepackage{booktabs}
\usepackage[caption=false,font=footnotesize,labelfont=sf,textfont=sf]{subfig}
\journalname{International Journal of Computer Vision}
\begin{document}\sloppy

\title{Real-world Video Deblurring: A Benchmark Dataset and An Efficient Recurrent Neural Network
}

\titlerunning{Real-world Video Deblurring}        

\author{Zhihang Zhong         \and
        Ye Gao\and Yinqiang Zheng\and Bo Zheng\and Imari Sato 
}


\institute{Zhihang Zhong \at
              The University of Tokyo, Japan
            \and
            Ye Gao \at
              Tokyo Research Center, Huawei, Japan
            \and
            Yinqiang Zheng \at
              The University of Tokyo, Japan\\
              \email{yqzheng@ai.u-tokyo.ac.jp}           
           \and
           Bo Zheng \at
              Tokyo Research Center, Huawei, Japan
           \and
           Imari Sato \at
              National Institute of Informatics, Japan
           \and
}

\date{Received: date / Accepted: date}

\maketitle

\begin{abstract}
Real-world video deblurring in real time still remains a challenging task due to the complexity of spatially and temporally varying blur itself and the requirement of low computational cost. To improve the network efficiency, we adopt residual dense blocks into RNN cells, so as to efficiently extract the spatial features of the current frame. Furthermore, a global spatio-temporal attention module is proposed to fuse the effective hierarchical features from past and future frames to help better deblur the current frame. Another issue that needs to be addressed urgently is the lack of a real-world benchmark dataset. Thus, we contribute a novel dataset (BSD) to the community, by collecting paired blurry/sharp video clips using a co-axis beam splitter acquisition system. Experimental results show that the proposed method (ESTRNN) can achieve better deblurring performance both quantitatively and qualitatively with less computational cost against state-of-the-art video deblurring methods. In addition, cross-validation experiments between datasets illustrate the high generality of BSD over the synthetic datasets. The code and dataset are released at \href{https://github.com/zzh-tech/ESTRNN}{https://github.com/zzh-tech/ESTRNN}.
\end{abstract}

\section{Introduction}
\label{sec:intro}

\begin{figure*}[!t]
	\centering
	\subfloat[]{\includegraphics[width=0.41\textwidth]{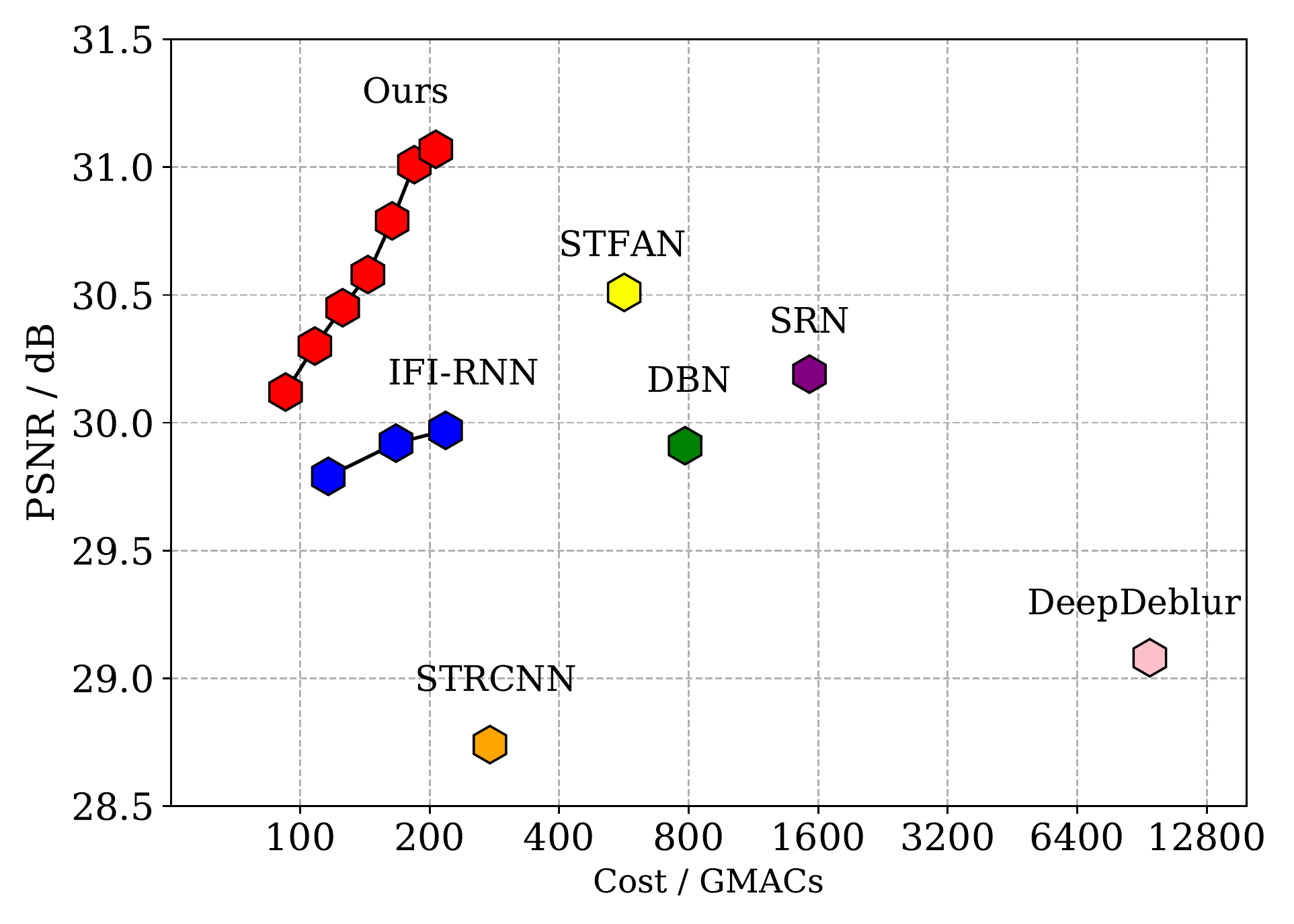}
		\label{fig:1a}}
	\hfil
	\subfloat[]{\includegraphics[width=.56\textwidth]{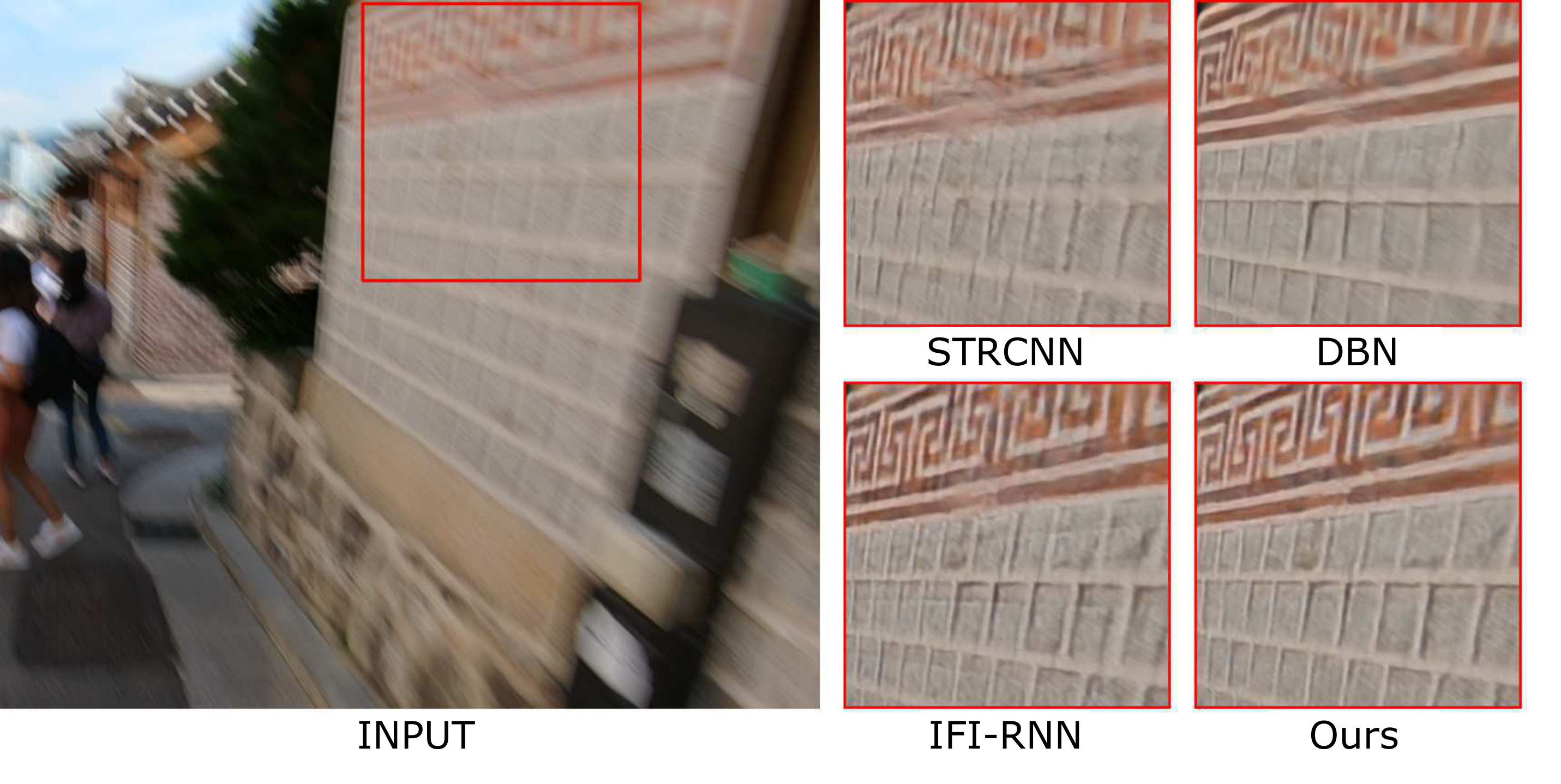}
		\label{fig:1b}}
	\caption{A comparison of network efficiency on video deblurring. SRN~\cite{tao2018scale}, DeepDeblur~\cite{nah2017deep} are methods for image deblurring, and STRCNN~\cite{hyun2017online}, DBN~\cite{su2017deep}, IFI-RNN~\cite{nah2019recurrent} are methods for video deblurring. (a) shows the computational cost required for processing a frame of 720P($1280\times720$) video and the corresponding performance of each model on GOPRO~\cite{nah2017deep} dataset in terms of GMACs and PSNR, respectively. (b) shows the deblurred image generated by SoTA video deblurring methods and ours.}
	\label{fig:efficiency}
\end{figure*}

Nowadays, video recording usually suffers from the quality issues caused by motion blur. This is especially true in poorly illuminated environment, where one has to lengthen the exposure time for sufficient brightness. A great variety of video deblurring methods have been proposed, which have to deal with two competing goals, i.e., to improve the deblurring quality and to reduce the computational cost. Reducing computational cost while ensuring deblurring quality is of critical importance for low-power mobile devices, such as smartphones.

To properly make use of the spatio-temporal correlation of the video signal is the key to achieve better performance on video deblurring. Deep learning-based methods have brought great advances in the field of video deblurring. The CNN-based methods~\cite{su2017deep,wang2019edvr} make an inference of the deblurred frame by stacking neighboring frames with current frame as input to the CNN framework. The RNN-based methods, like \cite{wieschollek2017learning,hyun2017online,zhou2019spatio,nah2019recurrent}, employ recurrent neural network architecture to transfer the effective information frame by frame for deblurring. However, how to utilize spatio-temporal dependency of video for deblurring more efficiently still needs to be explored. The CNN-based methods are usually cumbersome in dealing with spatio-temporal dependency of concatenated neighboring frames, and the existing RNN-based methods have limited capacity to transfer the effective information temporally. Thus, they either suffer from huge computational cost, or ineffectiveness of deblurring.

The quality of the benchmark dataset is also critical to the deblurring performance of the data-driven methods. It is worth noting that all mainstream deblurring datasets~\cite{nah2017deep, su2017deep, nah2019ntire} are synthesized by averaging high-fps videos. Unnatural data distributions and artifacts in synthetic blur will inevitably lead to poor generalization of the models that are trained on synthetic datasets, especially for real blurry images/videos. The community is currently in urgent need of a real-world deblurring dataset. However, obtaining the corresponding ground truth for the captured blurry image/video remains an unsolved challenge.

\begin{figure*}[!t]
	\centering
	\includegraphics[width=\linewidth]{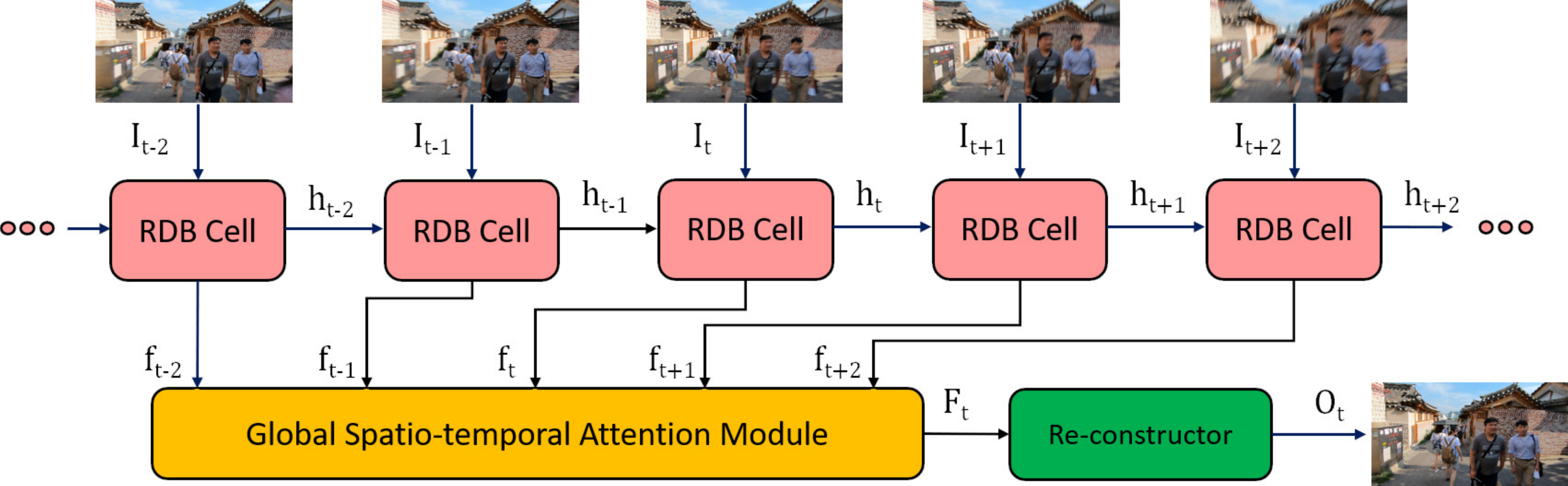}
	\caption{Framework of the proposed efficient spatio-temporal recurrent neural network. $I_t$ refers to the $t^{th}$ input blurry frame; $h_t$ and $f_t$ refer to the extracted hidden state and hierarchical features of RDB-based RNN cell (see Sec.~\ref{sec:RDB}) from $t^{th}$ frame; $F_t$ refers to the fused features generated by GSA module (see Sec.~\ref{sec:GSA}) for $t^{th}$ frame; $O_t$ refers to the $t^{th}$ deblurred frame by the proposed method.}
	\label{fig:process}
\end{figure*}

To optimize the trade-off between computational cost and deblurring performance, we propose an efficient spatio-temporal recurrent neural network, denoted by ESTRNN. In this work, we mainly focus on the network efficiency, which directly reflects on the deblurring performance of the method with limited computational resources. To make a more computationally efficient video deblurring method, we develop our method through amelioration of basic RNN architecture from three aspects: (i) In temporal domain, the high-level features generated by RNN cell are more informative, which are more suitable for temporal feature fusion (see Fig.~\ref{fig:process}) than using channel-concatenated neighboring frames as input. Reusing high-level features of neighboring frames can also help to improve the overall network efficiency; (ii) It is obvious that not all high-level features from neighboring frames are beneficial to deblurring of the current frame. Thus, it is worth designing an attention module~\cite{bahdanau2014neural} that allows the method to focus on more informative parts of the high-level features from other frames. To this end, we propose a novel global spatio-temporal attention module (see Sec.~\ref{sec:GSA}) for efficient temporal feature fusion; (iii) Regarding spatial domain, how to extract the spatial features from the current frame will affect the quality of information transmitted in temporal domain. In other words, well generated spatial features of each frame are a prerequisite for ensuring good temporal feature fusion. Therefore, we integrate the residual dense block~\cite{zhang2018residual} as backbone into RNN cell to construct RDB cell (see Sec.~\ref{sec:RDB}). The high-level hierarchical features generated by RDB cell are more computationally efficient with richer spatial information. With the above improvement, the proposed method can achieve better performance with less computational cost against SoTA deblurring methods, as illustrated in Fig.~\ref{fig:efficiency}(a). Due to making full use of spatio-temporal dependency of video signal, our method is exceptionally good at restoring high-frequency details of the blurry frame compared with SoTA video deblurring methods, as shown in Fig.~\ref{fig:efficiency}(b).

Furthermore, to get out of the predicament of lacking real-world deblurring dataset, we design a co-axis beam splitter acquisition system for data sample collection. Two cameras with distinct exposure schemes are co-axially aligned via a beam splitter to capture both blurry and sharp video of the same scene simultaneously. Empowered by the proposed beam-splitter acquisition system, we contribute the first real-world video deblurring dataset (BSD) to this field. BSD includes three different blur intensity configurations in dynamic scenes and contains various ego-motion and object-motion types. The advantages of the collected real-world dataset are verified by cross-validation experiments with synthetic datasets. Specifically, models trained on BSD can obtain migratory deblurring capabilities for other datasets; in contrast, the models trained on synthetic datasets have very limited generality, even on other synthetic datasets.  

Our major contributions can be summarized as follows:
\begin{itemize}
	\item To better exploit the spatio-temporal correlation of video signal for deblurring, we propose a novel RNN-based model that uses a recurrent cell based on residual dense blocks and a global spatio-temporal attention module to generate hierarchical spatial features and fuse the high-level features of neighboring frames, respectively.
	
	\item We design a beam splitter acquisition system to capture realistic blurry/sharp video pairs. To the best of our knowledge, the proposed BSD is the first real-world video deblurring dataset. 
	
	\item The experimental results demonstrate that our method achieves better deblurring performance both quantitatively and qualitatively than SoTA video deblurring methods with less computational cost.
	
	\item We have thoroughly done cross-validation between the proposed BSD and the synthetic datasets. The experimental results demonstrate that real-world dataset BSD outperforms the synthetic datasets in terms of generality of the trained models.
\end{itemize}

A short version of this paper was published in~\cite{zhong2020efficient}. Compared to~\cite{zhong2020efficient}, there are two main extensions: (i) We upgraded the beam-splitter acquisition system with center-aligned scheme and collected a larger real-world dataset for experiments with more scenes, more motion patterns and more blur intensity settings; (ii) We conducted cross-validation experiments between our real-world dataset BSD and the synthetic datasets, including existing dataset synthesized from \SI{240}{fps} videos and a self-made high-fps dataset synthesized from \SI{2000}{fps} videos, to validate the effectiveness of BSD.

\section{Related Works}
\label{sec:related}
\subsection{Video Deblurring}
\label{sec:related_deblur}
In recent years, video deblurring techniques become significant for daily life media editing and for advanced processing such as SLAM \cite{lee2011simultaneous}, 3D reconstruction \cite{seok2013dense} and visual tracking \cite{wu2011blurred}. Research focus starts to shift from early single non-blind image deblurring \cite{zoran2011learning, schuler2013machine, sun2014good} and single blind image deblurring \cite{xu2010two, goldstein2012blur, michaeli2014blind, chakrabarti2016neural, nimisha2017blur} to the more challenging tasks such as blur decomposition~\cite{jin2018learning,shen2020blurry,zhong2022animation}, and video deblurring~\cite{su2017deep,wang2019edvr,nah2019recurrent,zhou2019spatio,zhong2021towards,wang2022efficient}.

Typically, the blur in a video has varying size and magnitude at different positions in each frame. In the early work of video deblurring,~\cite{bar2007variational} attempts to automatically segment moving blurred objects from the background and assumes a uniform blur model for them. Then, in view of different kinds of blur in different regions of an image,~\cite{wulff2014modeling} tries to segment an image into two layers and generate segment-wise blur kernels for deblurring. More recently, there are some researches that estimate pixel-wise blur kernel with segmentation \cite{ren2017video}, or without segmentation \cite{hyun2014segmentation}. However, these kernel based methods are quite expensive in computation and usually rely on human knowledge. An inaccurate blur kernel estimation will result in severe artifacts in the deblurred image. Alternatively,~\cite{cho2012video} uses homography as the underlying motion model. Instead of deconvolution with kernel estimation, it searches for the luckier pixels in adjacent frames and uses them to replace the less lucky pixels in the current frame. However, the homography assumption can only be applied to the blur caused by camera shake.

To overcome the above issues, researchers started to work on deep learning methods for video deblurring. CNN-based methods are used to handle the inter-frame relationship of video signal, such as~\cite{su2017deep}, which makes the estimation of deblurred frame by using channel-concatenated neighboring frames with optional alignment using homography or optical flow. To better utilizing information from neighboring frames, EDVR~\cite{wang2019edvr} uses deformable convolutional operation~\cite{zhu2019deformable} in the encoder stage. Recently, on the basis of alignment using off-the-shelf optical flow estimators~\cite{sun2018pwc,hui2018liteflownet}, CDVD-TSP~\cite{pan2020cascaded} develops a temporal sharpness prior and a cascaded training approach to jointly optimize the network; while PVDNet~\cite{son2021recurrent} retrains a blur-invariant flow estimator and uses a pixel volume containing candidate sharp pixels to address motion estimation errors. However, additional flow estimator branch usually makes the model more computationally expensive. On the other hand, some researchers tend to focus on RNN-based methods because of their excellent performance for handling time-series signal. RNN-based methods can manage alignment implicitly through hidden states. For example,~\cite{wieschollek2017learning} employs RNN architecture to reuse the features extracted from the past frame, and~\cite{hyun2017online} improves the deblurring performance by blending the hidden states in temporal domain. Then,~\cite{nah2019recurrent} is proposed to iteratively update the hidden state via reusing RNN cell parameters and achieves impressive video deblurring performance while operating in real time.~\cite{zhou2019spatio} proposes a unified RNN framework to generate spatially adaptive filters for alignment and deblurring.~\cite{wang2022efficient} introduces a novel framework that utilizes the motion magnitude prior as guidance for efficient deep video deblurring. In addition,~\cite{zhong2021towards} considers the problem of joint rolling shutter correction and video deblurring in the case of using rolling shutter cameras.

In this paper, we adopt a RNN framework similar to~\cite{nah2019recurrent}. Our method is different from~\cite{nah2019recurrent} in that we integrate RDB into the RNN cell in order to exploit the potential of the RNN cell through feature reusing and generating hierarchical features for the current frame. Furthermore, we propose a GSA module to selectively merge effective hierarchical features from both past and future frames, which enables our model to utilize the spatio-temporal information more efficiently.

\subsection{Deblurring Dataset}
\label{sec:related_dataset}
As data-driven methods become dominant in the field of deblurring techniques, high-quality deblurring datasets with training image/video pairs become increasingly important. Since obtaining ground truth for real blurry image/video has been a challenge for a while, researchers thereby turned to synthetic datasets to circumvent this problem. A common approach used to create blurry images is to convolve clean natural images with synthetic blur kernels, just as was done in these works~\cite{xu2014deep, sun2015learning, chakrabarti2016neural, kupyn2018deblurgan}. A more typical approach is to average consecutive sharp frames in a high-fps video to create visually realistic blur that forms due to relatively long exposure, such as DVD~\cite{su2017deep}, GOPRO~\cite{nah2017deep} and REDS~\cite{nah2019ntire}. Research on the synthesis of blur on RAW space~\cite{cao2022towards} has also been carried out. However, the unnatural data distribution and artifacts in the synthetic deblurring datasets will inevitably affect the performance of the model in real-world situation.

Recently, researchers have resorted to designing customized hardware platforms to collect real-world dataset~\cite{zhang2019zoom,zhong2020efficient,rim2020real,zhong2021towards,cao2022learning} for low-level vision tasks. Take the image super-resolution task as an example, a zoom lens is used in~\cite{zhang2019zoom} to collect high-resolution and low-resolution image pairs in static scenes. In this work, we design a beam-splitter acquisition system for the video deblurring task and contribute the first real-world video deblurring dataset to the community. We found another work~\cite{rim2020real} that proposed a similar system to ours in the same period for collecting deblurring dataset. However, unlike~\cite{rim2020real}, we adopt the single object lens scheme to ensure precise alignment of the two cameras without receiving the effects of lens inter-reflection. Our carefully designed acquisition system is sufficient to capture blurry/sharp video pairs for both image and video deblurring, while the acquisition system in~\cite{rim2020real} can only be used to capture single blurry/sharp image pairs.

\section{Proposed Method}
\label{sec:method}
\begin{figure*}[!t]
	\centering
	\includegraphics[width=0.9\linewidth]{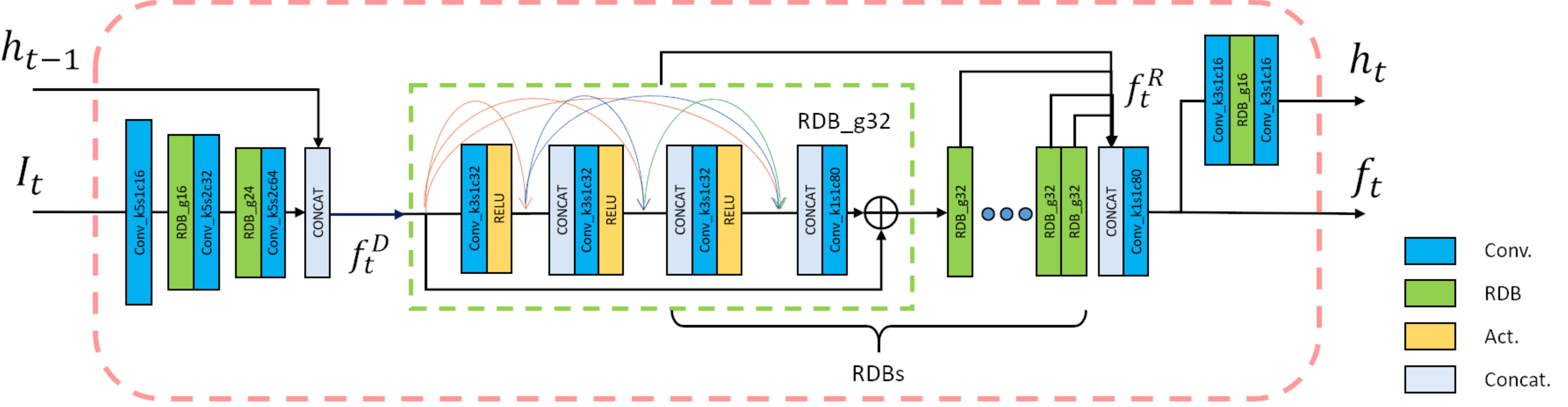}
	\caption{The structure of RDB-based RNN cell. $h_t$ and $h_{t-1}$ refer to the hidden state of past frame and current frame, respectively; $I_t$ refers to the input blurry frame; $f_t^D$ refers to the features after downsampling module; $f_t^R$ refers to the feature set generated by series of RDB modules; $f_t$ refers to the hierarchical features generated by the RDB cell; As for the details of each layer and RDB module, $k$, $s$, $c$ and $g$ denote kernel size, stride, channels and growth rate, respectively.}
	\label{fig:rdb}
\end{figure*}

\begin{figure}[!t]
	\centering
	\includegraphics[width=\linewidth]{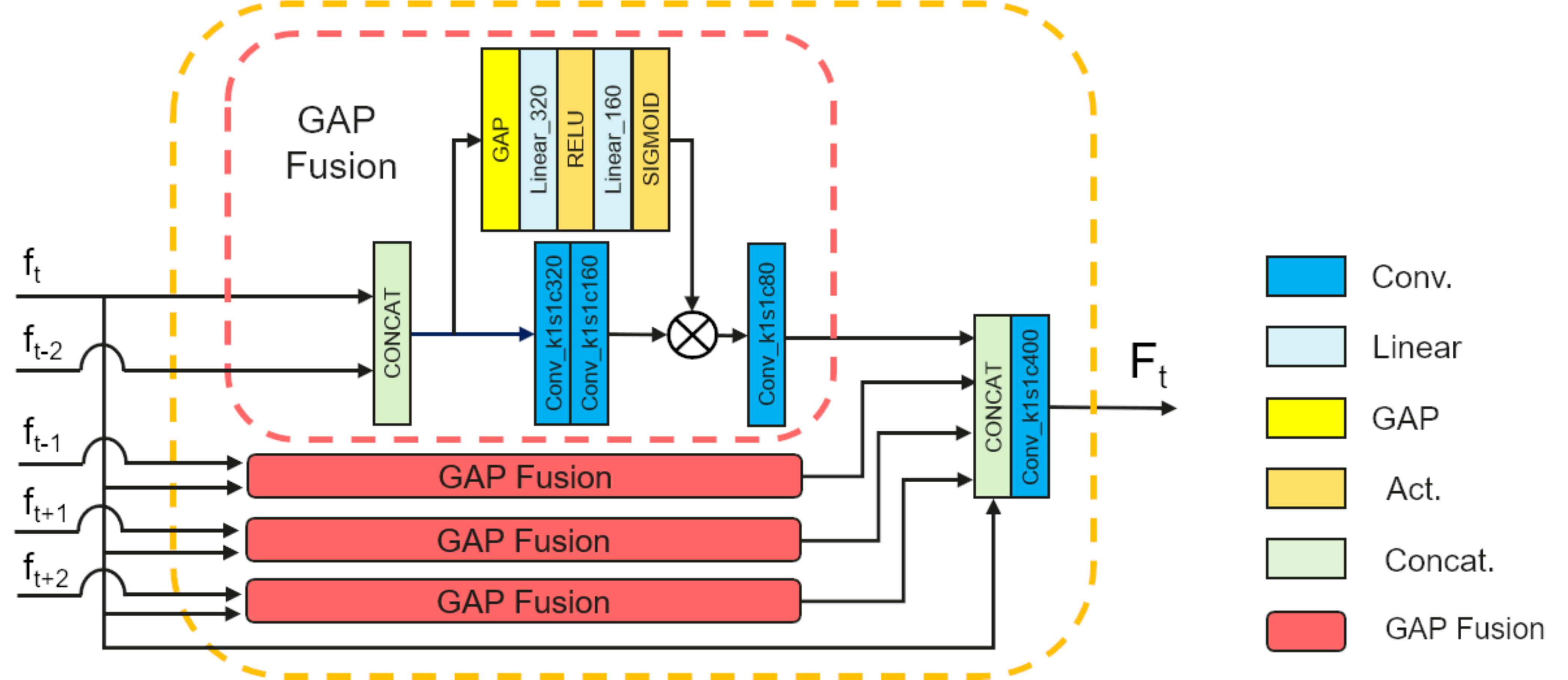}
	\caption{The structure of global spatio-temporal attention module. $f_{t-2}$, $f_{t-1}$, $f_{t+1}$, $f_{t+2}$ and $f_t$ refer to the hierarchical features of corresponding neighboring frames in the past and future and the current frame, respectively; $Linear$ refers to fully convolutional layer; $GAP$ refers to global average pooling fusion module; $F_t$ refers to the output of GSA module, integrating the effective components of hierarchical features from neighboring frames generated by $GAP$ fusion module.}
	\label{fig:gsa}
\end{figure}

In this section, we will first give an overview of the proposed method in Sec.~\ref{sec:overview}. Then we will go into details of RDB cell and GSA module in Sec.~\ref{sec:RDB} and Sec.~\ref{sec:GSA}, respectively.

\subsection{Overview}
\label{sec:overview}

According to the characteristics of blur in the video, it may keep varying temporally and spatially, which makes deblurring problem intractable. In turn, it is possible that the blurred information in the current frame is relatively clear and complete in the past frames and future frames. When using RNN-based method to implement video deblurring, high-level features of the current frame will be generated to make the inference of deblurred image. Actually, some parts of the high-level features are worth saving and reusing for making up the loss information for other frames. Therefore, distributing part of computational resources to fuse informative features in past and future frames could be a method to effectively improve the efficiency of the neural network. Furthermore, how to improve RNN cell itself to extract high-level features with better spatial structure is critical to enhance the efficiency of the neural network. Starting from the above viewpoints, we integrate multiple residual dense blocks into RNN cell to generate hierarchical features and propose a global spatio-temporal attention module for effective feature fusion of neighboring frames.

The whole video deblurring process of our method is shown as Fig.~\ref{fig:process}. We denote the input frames of blurry video and corresponding output frames as $\{I_t\}$ and $\{O_t\}$ respectively, where $t \in \{1\cdots T\}$. Through RDB-based RNN cell, the model could get hierarchical features for each frame as $\{f_t\}$. To get the inference of latent frame $O_t$, the global spatio-temporal attention module takes current hierarchical feature $f_t$ with two past and two future features $(f_{t-2}, f_{t-1}, f_{t+1}, f_{t+2})$ as input to perform feature fusion and generate $F_t$ as output. Finally, through re-constructor module, the model can obtain the latent frame $O_t$.

\subsection{RDB Cell: RDB-based RNN Cell}
\label{sec:RDB}

We adopt residual dense block \cite{zhang2018residual, zhang2020residual} into the RNN cell, named as RDB cell. The dense connections of RDB inherited from dense block (DB) \cite{huang2017densely} let each layer receive feature maps from all previous layers by concatenating them together in channel dimension. The output channels of each layer in RDB will keep the same size. This allows collective features to be reused and save the computational resources. Moreover, through local feature fusion, RDB could generate hierarchical features from convolutional layers in different depth with different size of receptive fields, which could provide better spatial information for image reconstruction.

The structure of RDB-based RNN cell is illustrated in Fig.~\ref{fig:rdb}. First, the current input frame $I_t$ will be downsampled and concatenated with last hidden state $h_{t-1}$ to get shallow feature maps $f^{D}_t$ as follows:
\begin{equation}
f^{D}_t = \mathit{CAT}(\mathit{DS}(I_t), h_{t-1}),
\end{equation}
where $\mathit{CAT(\cdot)}$ refers to concatenation operation; $\mathit{DS(\cdot)}$ refers to downsampling operation in the cell which consists of $5\times5$ convolutional layers and RDB module. Then, $f^{D}_t$ will be fed into a series of RDB modules. The collective outputs of stacked RDB modules are represented as $f^R_t=\{f^{R_1}_t, \cdots, f^{R_N}_t\}$, where $N$ refers to the number of RDB modules. RDB cell could obtain the global hierarchical features $f_t$ by fusing the concatenation of local hierarchical features $f^R_t$ with $1\times1$ convolutional layer as follows:
\begin{equation}
f_t=\mathit{Conv}(\mathit{CAT(f^R_t)}),
\end{equation}
where $\mathit{Conv(\cdot)}$ refers to convolutional operation. Then, the hidden state $h_t$ could be updated as follows:
\begin{equation}
h_t = H(f_t),
\end{equation}
where $H$ refers to the hidden state generation function, consisting of $3\times3$ convolutional layer and RDB module. In short, while processing each frame in the video, the inputs of RDB cell are current blurry frame and previous hidden state. Then, RDB cell will generate the hierarchical features of this frame and update the hidden state as well.

\begin{figure*}[!ht]
	\centering
	\subfloat[]{\includegraphics[width=0.44\textwidth]{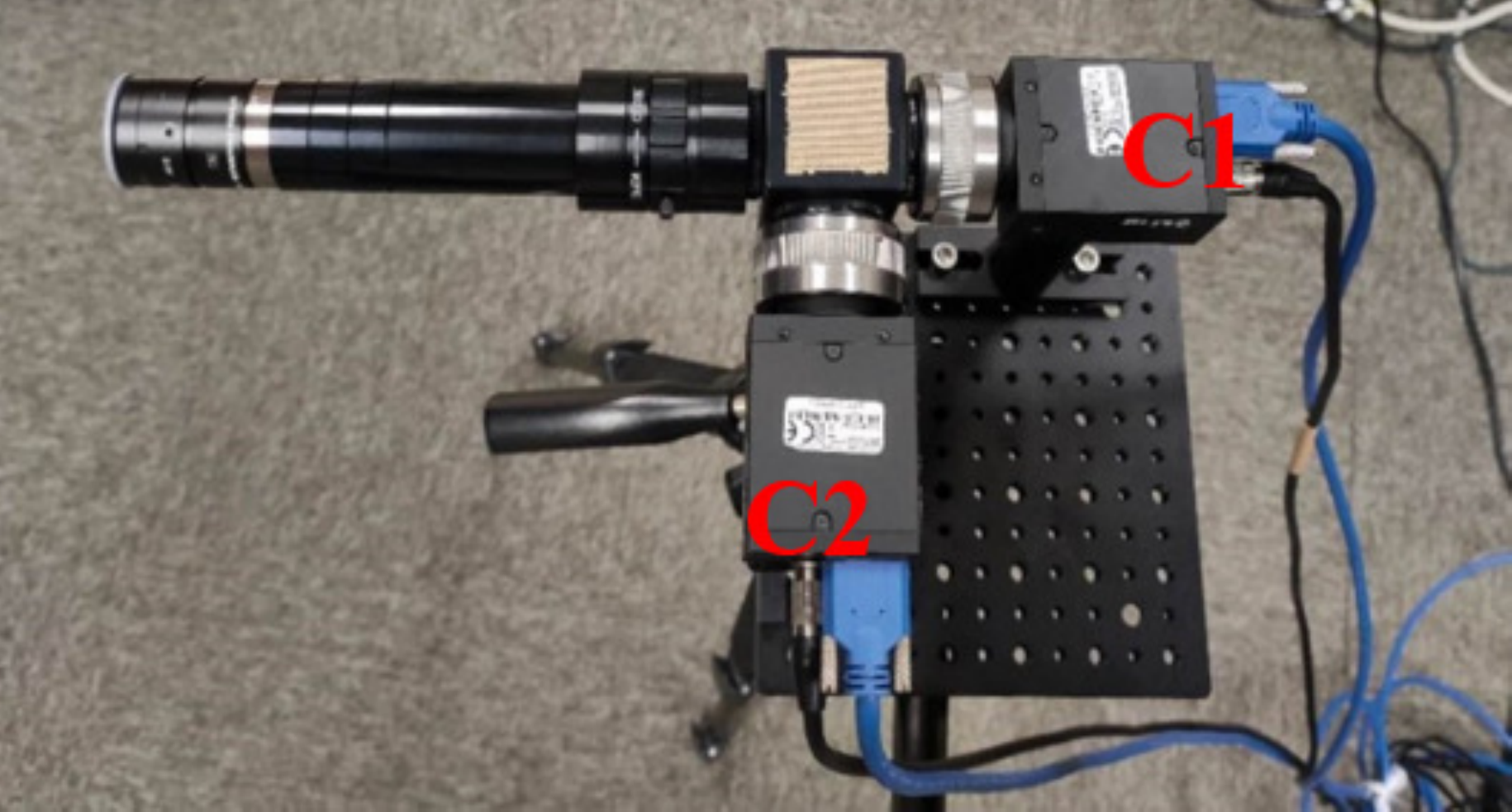}
		\label{fig:5a}}
	\hfil
	\subfloat[]{\includegraphics[width=.48\textwidth]{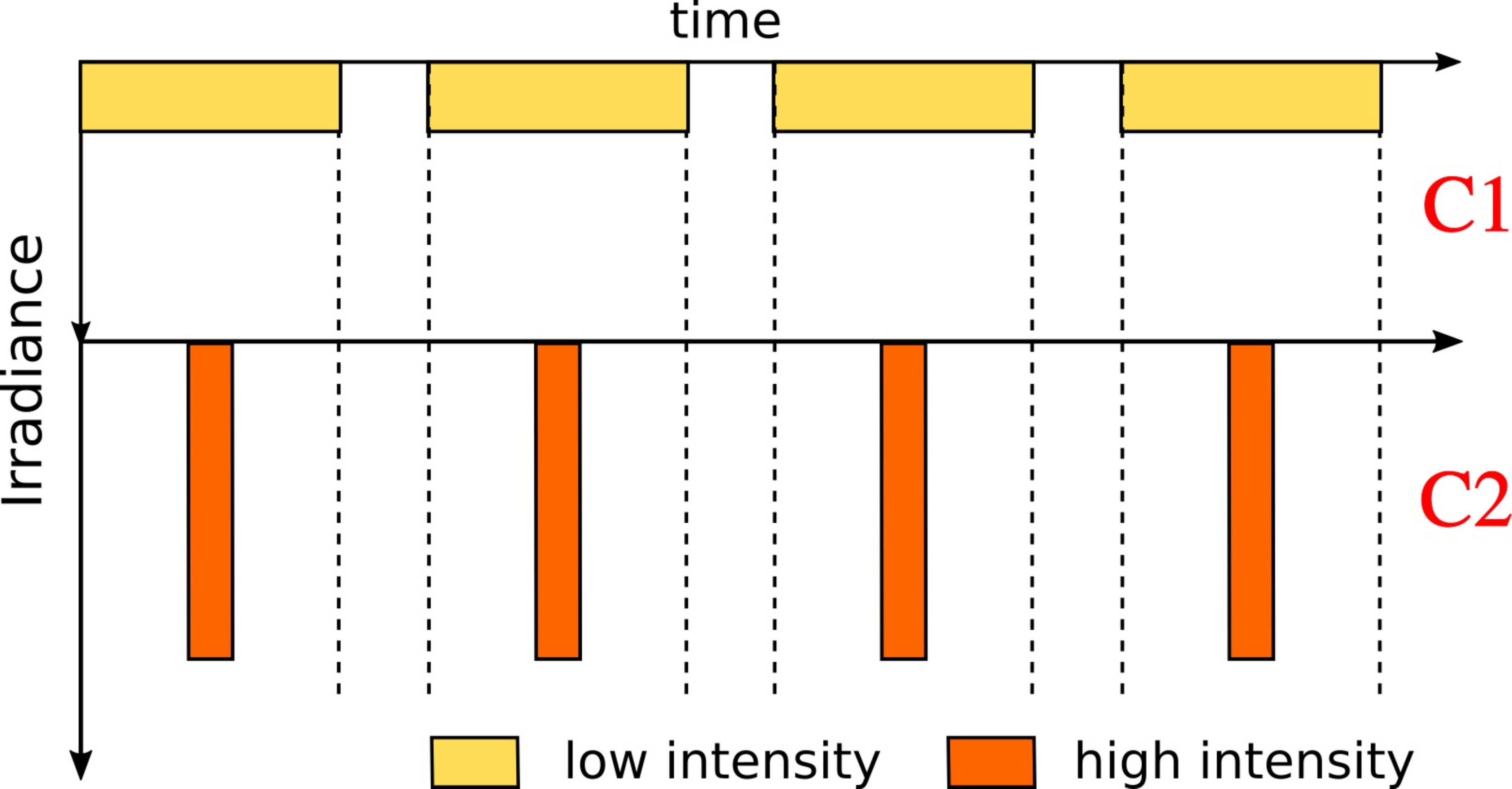}
		\label{fig:5b}}
	\caption{A beam splitter acquisition system for building video deblurring dataset. (a) is the profile of our beam splitter acquisition system. $C1$ and $C2$ refer to two same cameras with different configurations for generating blurry and sharp videos, respectively; (b) shows the center-aligned exposure scheme of $C1$ and $C2$ to generate blurry/sharp video pairs.}
	\label{fig:system}
\end{figure*}

\subsection{GSA: Global Spatio-temporal Attention Module}
\label{sec:GSA}

The structure of GSA module is illustrated in Fig.~\ref{fig:gsa}. This module aims to extract and fuse the effective components of hierarchical features from future and past frames. Intuitively, the frames which are closer to the current frame in time domain are more likely to have useful information for deblurring of current frame. In the situation of real-time video deblurring, considering that the requirement of low computational cost for each output frame, the number of neighboring hierarchical features that will be fused into current frame should be limited. Furthermore, considering that delaying output by only several frames is usually acceptable, the hierarchical features from the future frames are available for the feature fusion. Therefore, the input to GSA will be hierarchical features of two frames before and after the current frame and the current frame itself as $\{f_{t-2}, f_{t-1}, f_{t}, f_{t+1}, f_{t+2}\}$. Inspired by Squeeze-and-Excitation (SE) block in~\cite{hu2018squeeze}, a submodule named global averaging pooling fusion is proposed, which takes features of current frame and a neighboring frame as input to filter out effective hierarchical features $f^e_{t+i}$ from the neighboring frame as follows:
\begin{align}
f^{c}_{t+i} &= \mathit{CAT}(f_t, f_{t+i}),\\
f^{e}_{t+i} &= \mathit{L}(\mathit{GAP}(f^{c}_{t+i})) \otimes \mathit{P}(f^{c}_{t+i}), 
\end{align}
where $i \in \{-2,-1,1,2\}$; $\mathit{GAP}(\cdot)$ refers to global averaging pooling \cite{lin2013network}; $\mathit{L}(\cdot)$ refers to a series of linear transformation with activation function as $ReLU$~\cite{nair2010rectified} and $Sigmoid$ function for channel weight generation; $\mathit{P}(\cdot)$ refers to a series of $1\times1$ convolutional operations for feature fusion. Finally, GSA module will fuse the $f_t$ with all effective hierarchical features from neighboring frames to get the output $F_t$ as follows:
\begin{equation}
F_t = \mathit{Conv}(\mathit{CAT}(f^{e}_{t-2}, f^{e}_{t-1}, f^{e}_{t+1}, f^{e}_{t+2}, f_t)).
\end{equation}
The output $F_t$ of GSA module will be upsampled by deconvolutional layers~\cite{dumoulin2016guide} in re-constructor module for generating latent image for the current frame.

\section{Dataset for Video Deblurring}

\begin{figure*}[!ht]
	\centering
	\subfloat[]{\includegraphics[width=.24\textwidth]{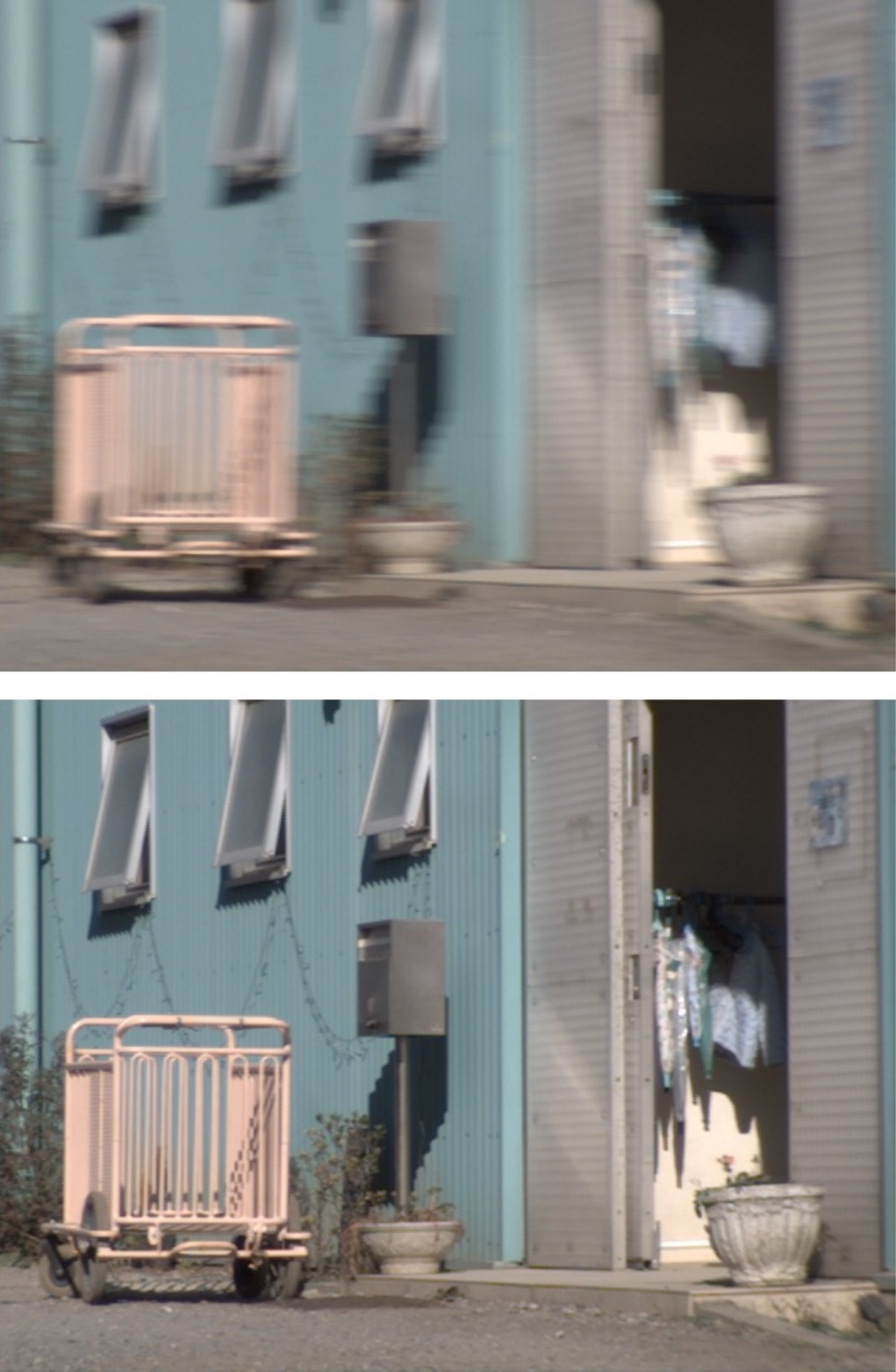}
		\label{fig:camera_move}}
	\hfil
	\subfloat[]{\includegraphics[width=.24\textwidth]{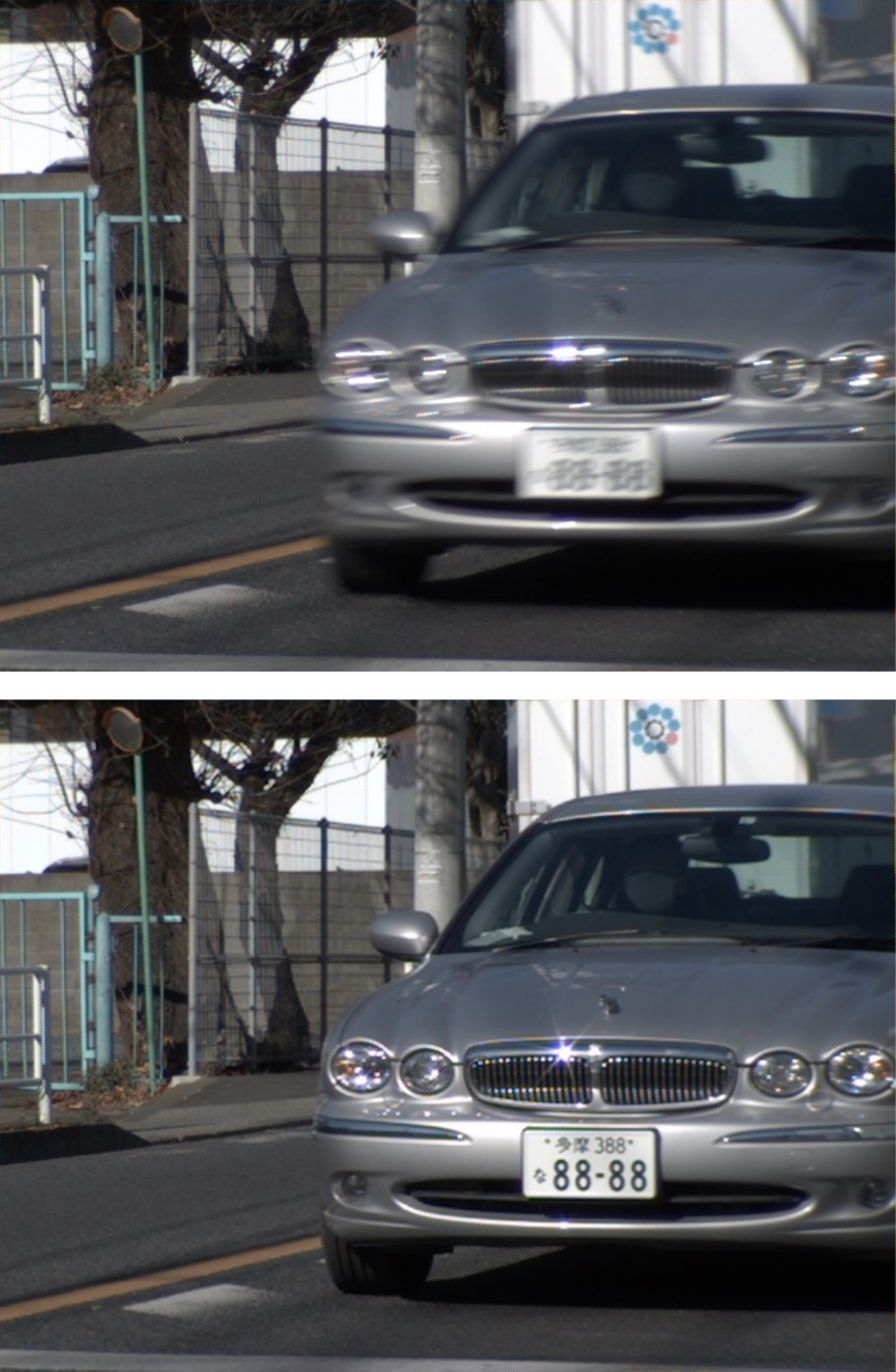}
		\label{fig:object_move}}
	\hfil
	\subfloat[]{\includegraphics[width=.24\textwidth]{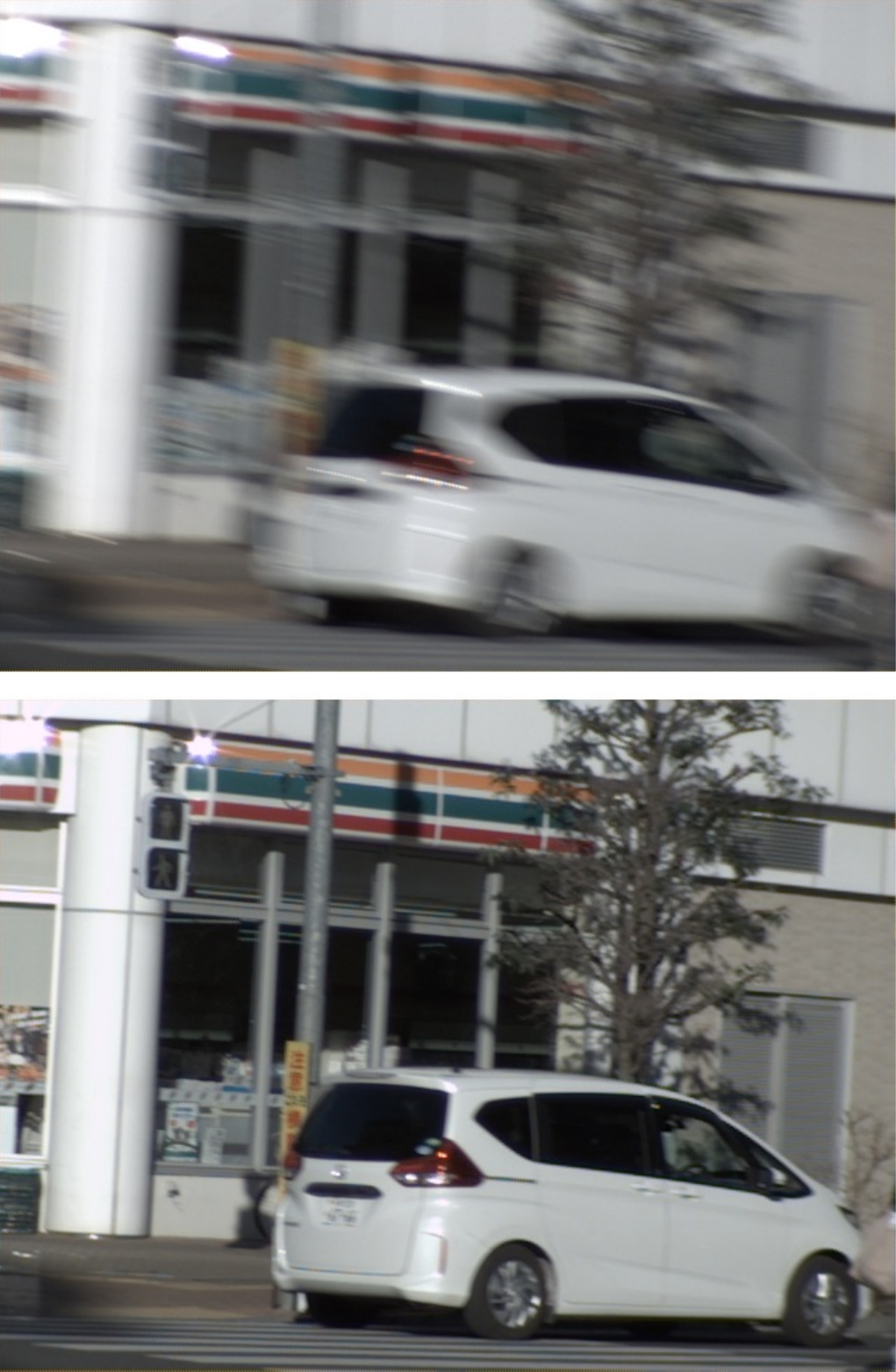}
		\label{fig:inverse_move}}
	\hfil
	\subfloat[]{\includegraphics[width=.24\textwidth]{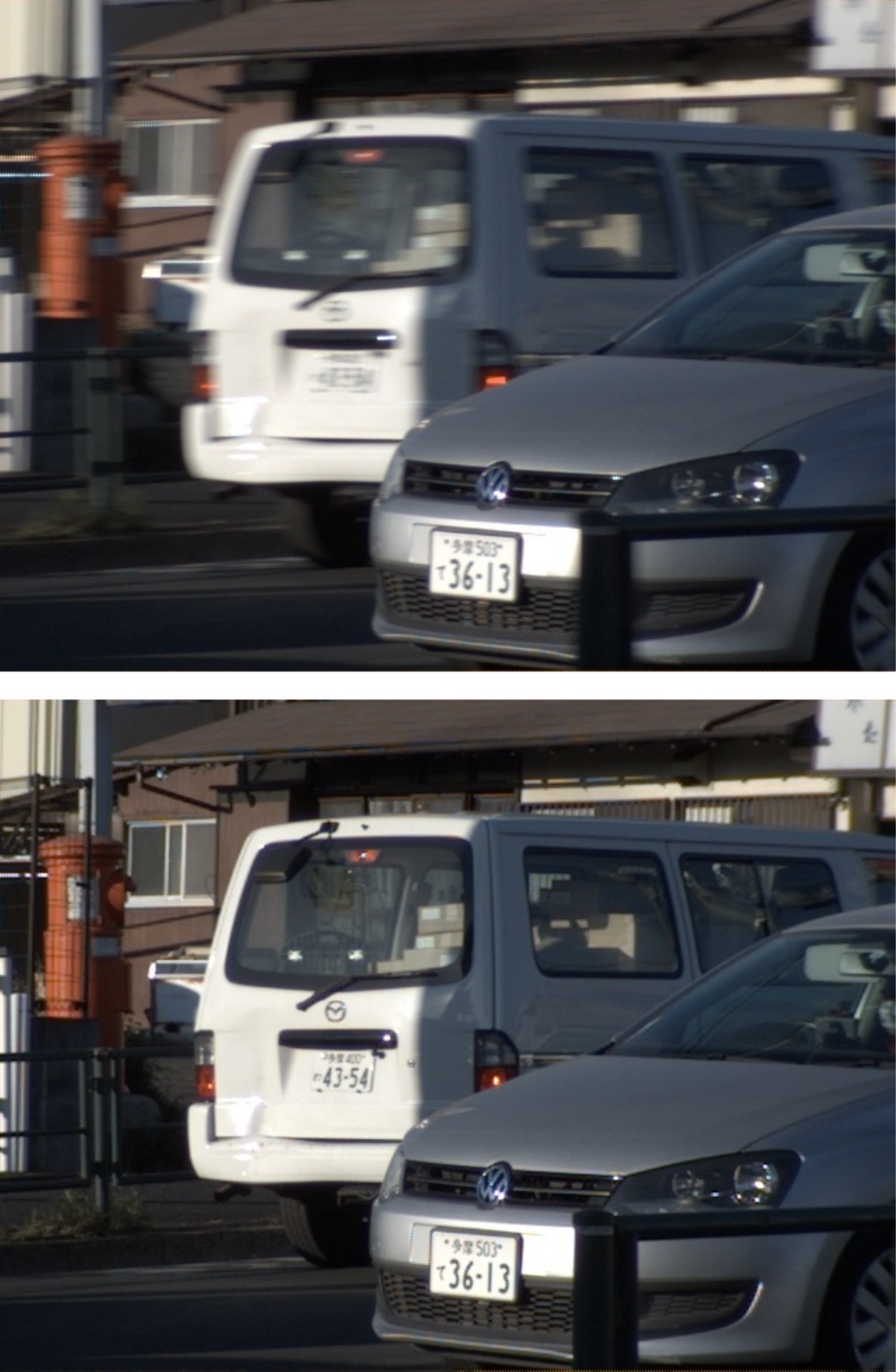}
		\label{fig:same_move}}
	\caption{Samples of blurry/sharp image pairs in BSD. (a) represents the case where there is only camera ego-motion. (b) represents the case where only the object is moving. (c) represents the case where the object and the camera are moving in the opposite directions. (d) represents the case where the object and the camera are moving in the same direction.}
	\label{fig:samples}
\end{figure*}

In this section, we first briefly introduce the mainstream synthetic video deblurring datasets and the corresponding simulation method. Then we present the details of our beam-splitter acquisition system and the corresponding BSD dataset.

\subsection{Synthesized Dataset}

At present, there are still very limited methods for building a video deblurring dataset. The mainstream way is to average a series of consecutive short-exposure images in order to mimic the phenomenon of blur caused by relatively long exposure time~\cite{kim2016dynamic}. The process of that can be described as follows:
\begin{equation}
	I_{b} = CRF\left( \frac{1}{N}\sum_{n=1}^{N}S^{n} \right),
\end{equation}
where $N$, $S^n$ denote the number of sampled high-FPS frames and the signal of $n^{th}$ sharp frame; $CRF$ denotes the camera response function.

This kind of method requires a high-speed camera to capture high-fps video and then synthesizes pairs of sharp and blurry videos based on the high-FPS videos. Before averaging, video frame interpolation algorithms~\cite{niklaus2017video} are usually used in advance to supplement the information of missing exposure time in the high-fps video, which helps to avoid unnatural spikes or steps in the blur trajectory caused by inadequate frame rate~\cite{wieschollek2017learning}. Mainstream public datasets for video deblurring, such as GOPRO \cite{nah2017deep} and REDS \cite{nah2019ntire}, are all born by the above method. There are 22 training video sequences and 11 evaluation video sequences in GOPRO with 2103 training samples and 1111 evaluation samples respectively. As for REDS, there are 240 training sequences and 30 evaluation sequences with 100 frames for each sequence.

\subsection{Beam-Splitter Deblurring Dataset (BSD)}

\begin{table}[b]
	\caption{Configuration of the proposed BSD dataset}
	\label{table:bsd_config}
	\begin{center}
		\begin{tabular}{lccc}
			\toprule
			&  train & validation & test\\
			\midrule
			sequences& $60$ & $20$ & $20$\\
			frames/seq.& $100$  & $100$ & $150$\\
			frames& $6000$  & $2000$ & $3000$\\
			resolution& \multicolumn{3}{c}{$640\times480$}\\
			frequency& \multicolumn{3}{c}{\SI{15}{fps}}\\
			settings& \multicolumn{3}{c}{1ms-8ms, 2ms-16ms, 3ms-24ms}\\
			\bottomrule
		\end{tabular}
	\end{center}
\end{table}

It is questionable whether aforementioned synthetic way truly reflects the blur in real scenarios. Here, we provide a new solution for building video deblurring dataset by using a beam splitter acquisition system with two synchronized cameras, as shown in Fig.~\ref{fig:system}(a). In our solution, by controlling the length of exposure time and strength of exposure intensity during video shooting as shown in Fig.~\ref{fig:system}(b), the system could obtain a pair of sharp and blurry videos in one shot. We adopt center-aligned synchronization scheme to properly delay the pulse of $C2$, so that the sharp exposure time lies exactly in the middle of the blurry exposure time. Compared to start-aligned or end-aligned synchronization scheme, center-aligned scheme can avoid large displacement between blurry/sharp image pairs. To further realize photometric alignment, we insert a $12.5\%$ neutral density filter in the front of $C1$ to reduce the irradiance intensity as $1/8$ of camera $C2$. Correspondingly, the exposure time of camera $C1$ is always kept as $8$ times of camera $C2$.

The configurations of the proposed BSD dataset is illustrated in Table~\ref{table:bsd_config}. We collected blurry/sharp video sequences for three different blur intensity settings (sharp exposure time -- blurry exposure time), 1ms--8ms, 2ms--16ms and 3ms--24ms, respectively. The acquisition frequency is 15fps. For each setting, the training and validation sets have 60 and 20 video sequences with 100 frames in each, respectively, and the test set has 20 video sequences with 150 frames in each. There are 11000 blurry/sharp image pairs in total for each setting. The resolution of all videos is uniformly $640\times480$. Blurry/sharp image pairs of different motion patterns are illustrated in Fig.~\ref{fig:samples}. The $1^{st}$ row shows blurry images, and the $2^{nd}$ row shows the corresponding sharp images. Specifically, Fig.~\ref{fig:samples}(a) represents the case where only the camera is moving; Fig.~\ref{fig:samples}(b) represents the case where only the object is moving while the camera is static. Thus, the background is sharp but the car is blurry; Fig.~\ref{fig:samples}(c) represents the case where the camera and the object are moving in the opposite directions; Fig.~\ref{fig:samples}(d) represents the case where the camera and the object are moving in the same direction. As a result, the car appears sharp while the background is blurry.

\section{Experiments and Results}
\subsection{Implementation Details}

For comparison on synthetic datasets, we conduct the experiments on GOPRO and REDS, respectively. As for GOPRO, we choose the same version as~\cite{nah2019recurrent}. Due to the huge size of REDS dataset and limited computational resources, we train our model and other SoTA models only on first-half training sequences of REDS (120 sequences) for comparison. We train the model for 500 epochs by ADAM optimizer~\cite{kingma2014adam} ($\beta_1=0.9, \beta_2=0.999$) with initial learning rate as $10^{-4}$ (decay rate as 0.5, decay step as 200 epochs). We use RGB patches of $256\times256$ size in subsequence of $10$ frames as input to train the models. Also, we implement horizontal and vertical flipping for each subsequence as data augmentation. Mini-batch size is set to 4. The loss function is defined as MSE loss $\mathcal{L}_{MSE}$ as follows:
\begin{equation}
    \mathcal{L}_{MSE} = \frac{1}{TCHW}\sum_{t=1}^{T} \left\Vert O_t - O_t^{GT} \right\Vert_{2}^{2},
\end{equation}
where $T$, $C$, $H$, $W$ denote the number of frames and the number of channel, height, width for each frame; $O_t^{GT}$ refers to the ground truth of the $t^{th}$ frame.

Whereas, in the experiments on the proposed real-world dataset BSD, we adopt cosine annealing schedule to adjust learning rate with default learning rate as $3\times10^{-4}$. Mini-batch size is set to $8$. Length of subsequence is $8$. We use charbonnier loss function $\mathcal{L}_{char}$ ($\epsilon=1\times10^{-3}$) as follows:
\begin{equation}
    \mathcal{L}_{char} = \frac{1}{TCHW}\sum_{t=1}^{T} \left\Vert \sqrt{\left(O_t - O_t^{GT}\right)^2 + \epsilon^2} \right\Vert.
\end{equation}

\begin{table}[!ht]
	\caption{Quantitative results on both GOPRO and REDS datasets. Cost refers to the computational cost of the model for deblurring one frame of HD (720P) video in terms of GMACs. The meaning of cost is same for other tables and figures in this paper. For our model, $B_{\#}$ and $C_{\#}$ denote the number of RDB blocks in RDB cell and the number of channels for each RDB block, respectively.}
	\label{table:psnr}
	\centering
	\resizebox{0.48\textwidth}{28mm}{
	\begin{tabular}{lccccc}
		\toprule
			\multirow{2}{*}{Model}&\multicolumn{2}{c}{GOPRO}&\multicolumn{2}{c}{REDS}&\multirow{2}{*}{Cost}\\
		& PSNR & SSIM & PSNR & SSIM & \\
		\midrule
		STRCNN\cite{hyun2017online} & 28.74 & 0.8465 & 30.23 & 0.8708 & 276.20\\
		DBN\cite{su2017deep} & 29.91 & 0.8823 & 31.55 & 0.8960 & 784.75\\
		IFI-RNN ($c2h1$)\cite{nah2019recurrent} & 29.79 & 0.8817 & 31.29 & 0.8913 & 116.29\\
		IFI-RNN ($c2h2$)\cite{nah2019recurrent} & 29.92 & 0.8838 & 31.35 & 0.8929 & 167.09\\
		IFI-RNN ($c2h3$)\cite{nah2019recurrent} & 29.97 & 0.8859 & 31.36 & 0.8942 & 217.89\\
		STFAN\cite{zhou2019spatio} & 30.51 & 0.9054 & 32.03 & 0.9024 & 566.61\\
		CDVD-TSP~\cite{pan2020cascaded} & 30.94 & 0.9153 & 32.57 & 0.9161 & 5211.28\\
		PVDNet~\cite{son2021recurrent} & 32.13 & 0.9322 & 33.92 & 0.9322 & 1754.90\\
		ESTRNN ($B_9C_{60}$) & 30.12 & 0.8837 & 31.64 & 0.8930 & 92.57\\
		ESTRNN ($B_9C_{65}$) & 30.30 & 0.8892 & 31.63 & 0.8965 & 108.20\\
		ESTRNN ($B_9C_{70}$) & 30.45 & 0.8909 & 31.94 & 0.8968 & 125.55\\
		ESTRNN ($B_9C_{75}$) & 30.58 & 0.8923 & 32.06 & 0.9022 & 143.71\\
		ESTRNN ($B_9C_{80}$) & 30.79 & 0.9016 & 32.33 & 0.9060 & 163.61\\
		ESTRNN ($B_9C_{85}$) & 31.01 & 0.9013 & 32.34 & 0.9074 & 184.25\\
		ESTRNN ($B_9C_{90}$) & 31.07 & 0.9023 & 32.63 & 0.9110 & 206.70\\
		\bottomrule
	\end{tabular}
	}
\end{table}

\subsection{Experiments on Synthetic Datasets}

For fair comparison, we mainly compare our method with lightweight deblurring methods without using pretrained optical flow estimator. We note that the SoTA models, such as CDVD-TSP and PVDNet, use a pre-trained optical flow estimator (PWC-Net~\cite{sun2018pwc} for CDVD-TSP and LiteFlowNet~\cite{hui2018liteflownet} for PVDNet) as part of the model, which means additional information is introduced into the model. Yet we also present the comparison results with CDVD-TSP~\cite{pan2020cascaded} and PVDNet~\cite{son2021recurrent} for your reference.

\subsubsection{Comparison on GOPRO}
\begin{figure*}[!t]
	\centering
	\includegraphics[width=\linewidth]{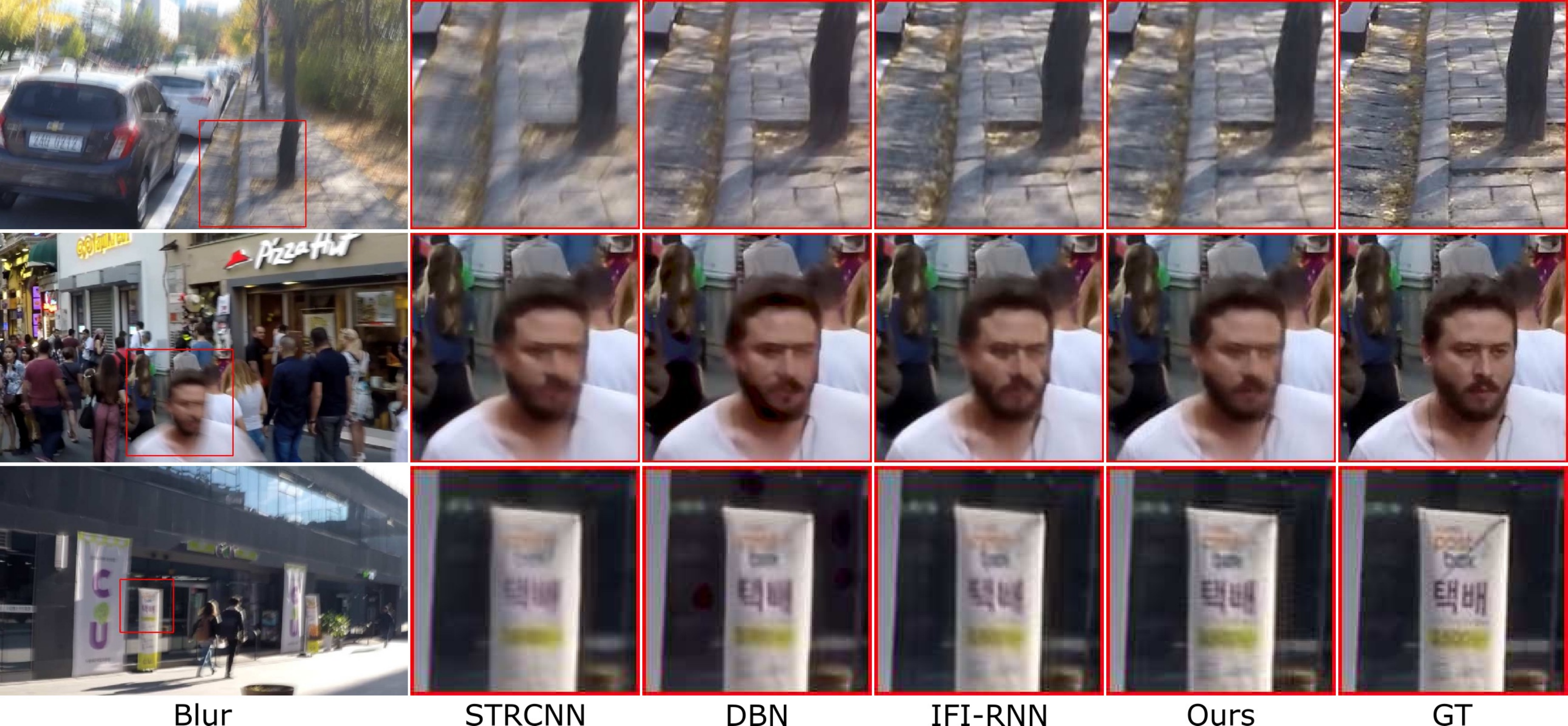}
	\caption{Visual comparisons on GOPRO~\cite{nah2017deep}.}
	\label{fig:gopro_results}
\end{figure*}

First, we compare our method with the SoTA video deblurring methods on GOPRO dataset. We implement 7 variants of our model with different computational cost by modifying the number of channels ($C_{\#}$) of the model and keeping the number of RDB blocks ($B_{\#}$) as $9$. The larger $C_{\#}$ is, the higher computational cost it needs. We report the deblurring performance and the corresponding computational cost for processing one frame in the video of all compared models in terms of PSNR \cite{hore2010image}, SSIM and GMACs, respectively, in Table~\ref{table:psnr}. From the perspective of quantitative analysis, it is clear that our model can achieve higher PSNR and SSIM value with less computational cost, which means our model has higher network efficiency. To further validate the deblurring performance of proposed model, we also show the deblurred image generated by each model, as illustrated in Fig.~\ref{fig:gopro_results}. We can see that the proposed model can restore sharper image with more details, such as the textures of tiles on the path and the characters on the poster.

\subsubsection{Comparison on REDS}

\begin{figure*}[!t]
	\centering
	\includegraphics[width=\linewidth]{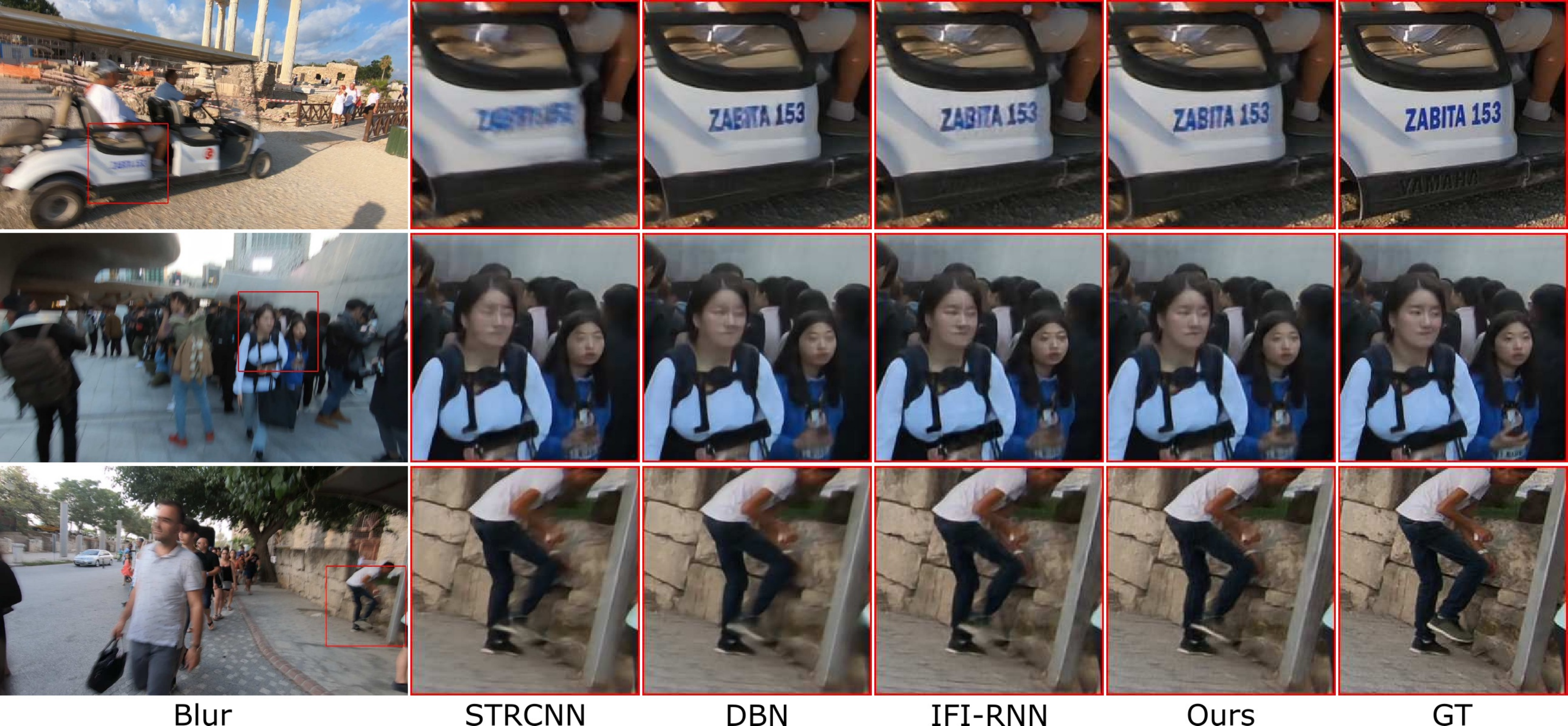}
	\caption{Visual comparisons on REDS~\cite{nah2019ntire}.}
	\label{fig:reds_results}
\end{figure*}

We also do the comparison on REDS, which has more diverse scenes from different places. From Table~\ref{table:psnr}, we can see our model $B_9C_{90}$ achieves best results as 32.63 PSNR with only around 200 GMACs computational cost for one $720P$ frame. Even our small model $B_9C_{60}$ with cost less than 100 GMACs can achieve same level performance as $c2h3$ of IFI-RNN, the computational cost of which is as twice as the former. In terms of qualitative results illustrated in Fig.~\ref{fig:reds_results}, the proposed model can significantly reduce ambiguous parts for the deblurred frame, and the restored details such as the texture of the wall, characters, and human body are closer to the ground truth.

\subsubsection{Network Efficiency Analysis}
\begin{table*}[ht]
  \caption{Computational cost and inference time comparison. We show the performance of each model in terms of GMACs, Million parameters, inference time (ms), frame rate (fps), and the corresponding PSNR score. The size of test image is 1280$\times$720, and the test hardware is GeForce RTX 2080 Ti. The average value is reported.}
  \label{tab:cost_comparison}
  \centering
  \begin{tabular}{lccccc}
    \toprule
    & STRCNN~\cite{hyun2017online} & DBN~\cite{su2017deep} & IFI-RNN~\cite{nah2019recurrent} & ESTRNN (B9C60) & ESTRNN (B15C80)\\
    \midrule
    GMACs & 276.20 & 784.75 & 217.89 & 92.57 & 203.74\\
    M Parameters& 0.93 & 15.31 & 1.64 & 0.99 & 2.47\\
    Inference Time (ms) & 42.03 & 61.9 & 67.0 & 33.7 & 62.2\\
    Frame rate (fps) & 23.79 & 16.16 & 14.93 & 29.66 & 16.07\\
    PSNR & 28.74 & 29.91 & 29.97 & 30.12 & 31.27 \\
    \bottomrule
  \end{tabular}
\end{table*}

We collect the computational cost for one frame as well as the performance (PSNR) of the SoTA lightweight image and video deblurring models on GOPRO dataset, as illustrated in Fig.~\ref{fig:efficiency}(b). The proposed model includes 7 red nodes that represent different variants of our ESTRNN from $B_9C_{60}$ to $B_9C{90}$ in Table~\ref{table:psnr}. Also, the three blue nodes represent different variants of IFI-RNN as $c2h1$, $c2h2$ and $c2h3$. Because the computational cost of different models varies drastically, we take $log_{10}(\text{GMACs})$ as abscissa unit to better display the results. An ideal model with high network efficiency will locate at upper-left corner of the coordinate. The proposed models are closer to the upper-left corner than the existing image or video deblurring models, which reflects the high network efficiency of our model.

We further present computational cost and inference time comparison in Table~\ref{tab:cost_comparison}. Because the parallelism of RNNs is in general worse than CNNs, RNN-based models do not have a considerable advantage in inference time, although they have a greater advantage in GMACs. However, our model still demonstrates efficiency improvements. Take the small variant ESTRNN (B9C60) as example, it is superior to other models in terms of performance, inference speed and computational cost. We also note that the parallelism of RNNs can be improved according to related literature~\cite{lei2017simple,yu2018sliced}.

\subsubsection{Ablation Study}

\begin{table}[!ht]
	\caption{Ablation study of ESTRNN. Fusion refers to the fusion strategy that utilizes the high level features from neighboring frames. The experiments are conducted on GOPRO.}
	\label{table:ablation}
	\centering
	\begin{tabular}{lccccc}
		\toprule
		Model & Fusion & RDB Cell & GSA & PSNR & Cost\\
		\midrule
		$B_9C_{110}$ & $\times$ & $\times$ & $\times$ & 30.29 & 163.48 \\
		$B_9C_{100}$ & \checkmark & $\times$ & $\times$ & 30.46 & 165.59 \\
		$B_9C_{100}$ & $\times$ & \checkmark & $\times$ & 30.51 & 168.56 \\
		$B_9C_{90}$ & \checkmark & \checkmark & $\times$ & 30.55 & 161.28 \\
		$B_9C_{85}$ & $\times$ & \checkmark & \checkmark & 30.69 & 162.69 \\
		$B_9C_{80}$ & \checkmark & \checkmark & \checkmark & 30.79 & 163.61 \\
		\bottomrule
	\end{tabular}
\end{table}

\begin{table}[!ht]
	\caption{Effectiveness of number of RDB blocks. The experiments are conducted on GOPRO.}
	\label{table:blocks}
	\centering
	\begin{tabular}{lccccc}
		\toprule
		& $B_3C_{80}$ & $B_6C_{80}$ & $B_9C_{80}$ & $B_{12}C_{80}$ & $B_{15}C_{80}$\\
		\midrule
		PSNR & 29.74 & 30.31 & 30.79 & 31.03 & 31.27 \\
		Cost & 123.03 & 143.32 & 163.31 & 183.90 & 204.19 \\
		\bottomrule
	\end{tabular}
\end{table}

\begin{table}[!ht]
	\caption{Effectiveness of number of neighboring frames used by GSA module. $F_{\#}$ and $P_{\#}$ refers to the number of future and past frames used by the model. The base model is $B_9C_{80}$. The experiments are conducted on GOPRO.}
	\label{table:frames}
	\centering
	\resizebox{0.48\textwidth}{8mm}{
	\begin{tabular}{lcccccc}
		\toprule
		& $F_0P_{1}$ & $F_0P_{2}$ & $F_0P_{3}$ & $F_{1}P_{1}$ & $F_{2}P_{2}$ & $F_{3}P_{3}$\\
		\midrule
		PSNR & 30.54 & 30.57 & 30.69 & 30.58 & 30.79 & 30.82 \\
		Cost & 119.93 & 133.75 & 148.31 & 133.75 & 163.61 & 196.42 \\
		\bottomrule
	\end{tabular}
	}
\end{table}

\begin{figure}[ht]
	\centering
	\includegraphics[width=0.48\textwidth]{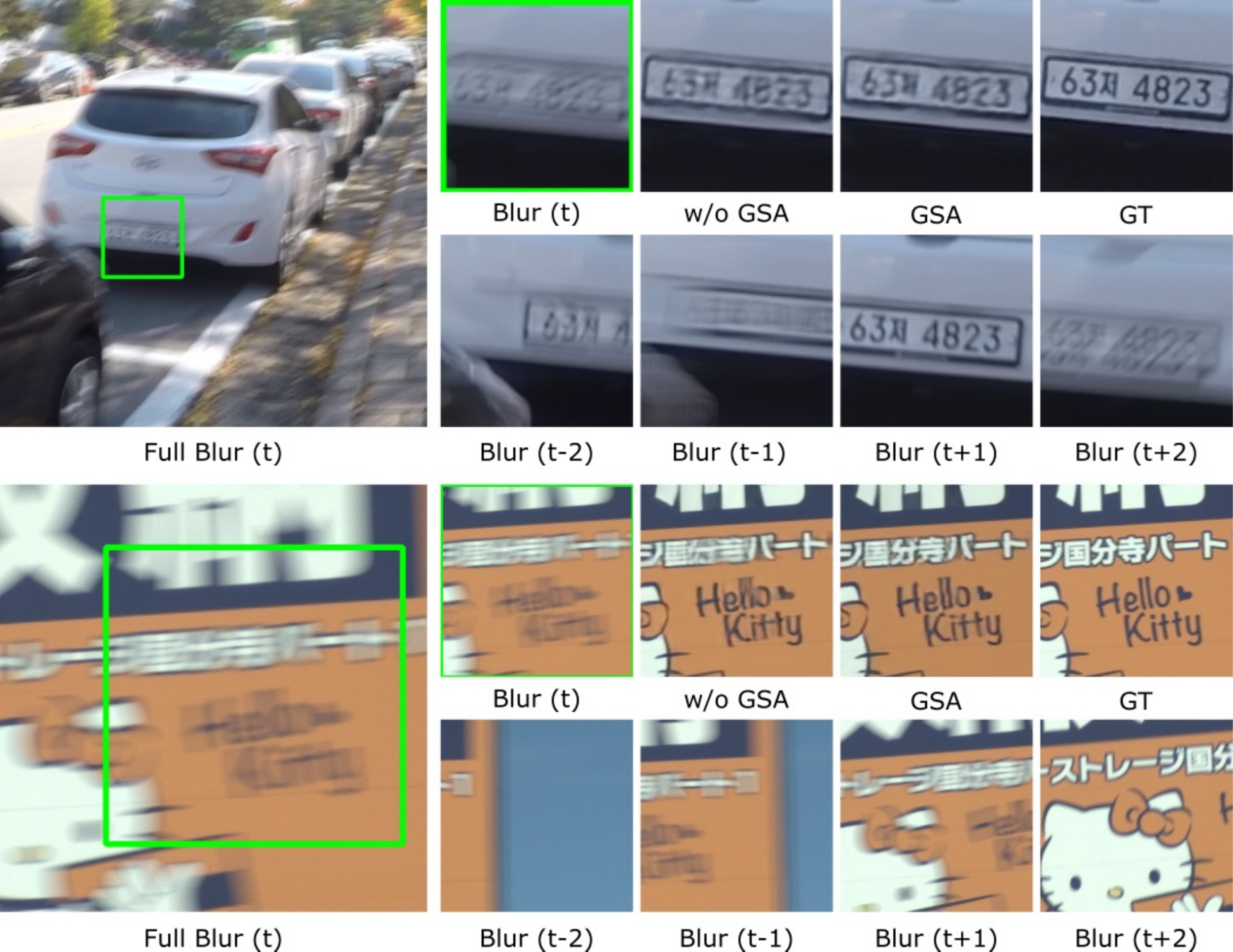}
	\caption{Visual ablation study of the GSA module.}
	\label{fig:visual_gsa}
\end{figure}

\begin{table*}[t]
	\caption{Quantitative results on BSD dataset. The configuration of ESTRNN is $B15C80$. The configuration of IFI-RNN is $C2H3$.}
	\label{table:bsd}
	\centering
	\setlength{\tabcolsep}{14pt}
	\begin{tabular}{lcccccc}
		\toprule
		\multirow{2}{*}{Model}&\multicolumn{2}{c}{1ms--8ms}&\multicolumn{2}{c}{2ms--16ms}&\multicolumn{2}{c}{3ms--24ms}\\
		& PSNR & SSIM & PSNR & SSIM & PSNR & SSIM\\
		\midrule
		STRCNN~\cite{hyun2017online} &32.20&  0.924&  30.33& 0.902& 29.42& 0.893 \\
		DBN~\cite{su2017deep} &33.22& 0.935& 31.75& 0.922& 31.21& 0.922 \\
		SRN~\cite{tao2018scale} &31.84  &0.917  &29.95 &0.891 & 28.92& 0.882\\
		IFI-RNN~\cite{nah2019recurrent} & 33.00&  0.933&  31.53& 0.919& 30.89& 0.917\\
		STFAN~\cite{zhou2019spatio} & 32.78 &  0.922&  32.19& 0.919& 29.47&0.872\\
		CDVD-TSP~\cite{pan2020cascaded} & 33.54& 0.942&  32.16&0.926 & 31.58&0.926\\
	    PVDNet~\cite{son2021recurrent} & 33.34&0.937&  32.22&0.926& 31.35&0.923\\
		ESTRNN &  33.36& 0.937&  31.95& 0.925& 31.39& 0.926\\
		\bottomrule
	\end{tabular}
\end{table*}

We conduct an ablation study to demonstrate the effectiveness of the high-level feature fusion strategy, RDB cell, as well as GSA module, as shown in Table~\ref{table:ablation}. When ablating the modules, we keep the computational cost almost unchanged by adjusting the number of channels ($C_{\#}$) for fair comparison. Specifically, without using fusion strategy means that the model directly reconstructs the result according to high-level features only from current frame; without RDB cell, the model will use residual block~\cite{he2016deep} instead, in the same way as~\cite{nah2017deep} does; without GSA module, high-level features will be directly concatenated in channel dimension. The results clearly demonstrate that each module or design can improve the deblurring efficiency, because each module can improve the overall performance of model when the computational cost keeps unchanged. Besides, we show visual results to provide some intuitions about GSA module for high-level feature fusion in Fig.~9. In these two cases, the results of model without using GSA to fuse features from neighboring frames (only features from current frame as input for the re-constructor) are inferior to the results of the complete model. The gain should come from the time instances, such as $t+1$ in the first case and $t+2$ in the second case, where the ``blurry'' input image has relatively sharp but misaligned appearance.

We further explore the effectiveness of the number of RDB blocks and the number of past and future frames used by the model as Table~\ref{table:blocks} and Table~\ref{table:frames}, respectively. First, from the perspective of the number of RDB blocks, this is intuitive that more blocks which means more computational cost will achieve better performance. If we compare the variant $B_{15}C_{80}$ with variant $B_9C_{90}$ in Table~\ref{table:psnr} which has almost same computational cost, we can find that it is better to increase the number of RDB blocks rather than the channels, when the number of channels is relatively sufficient. As for the number of neighboring frames, Table~\ref{table:frames} shows that, considering the increased computational cost, the benefit of using more neighboring frames as $F_3P_3$ is relatively small. Besides, the results of $F_0P_1$, $F_0P_2$ and $F_0P_3$ show that the proposed model can still achieve comparative good results even without high-level features from the future frames.

\subsection{Experiments on Real-world Dataset BSD}

\begin{figure*}[!ht]
	\centering
	\subfloat[Visual results on BSD (1ms--8ms)]{\includegraphics[width=\textwidth]{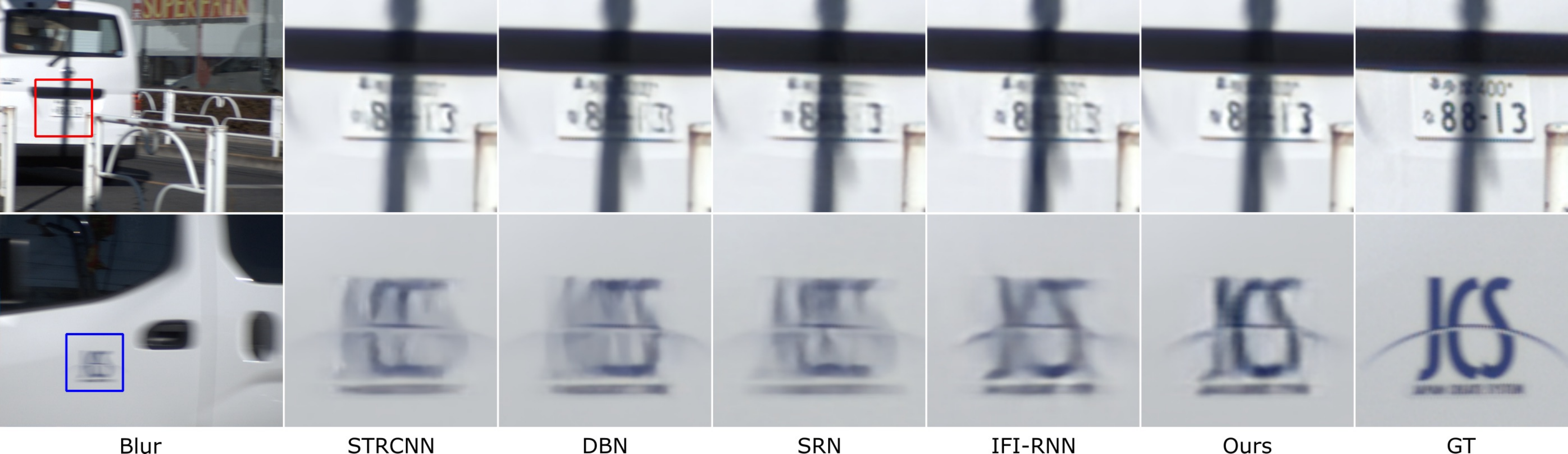}}
	\hfil
	\subfloat[Visual results on BSD (2ms--16ms)]{\includegraphics[width=\textwidth]{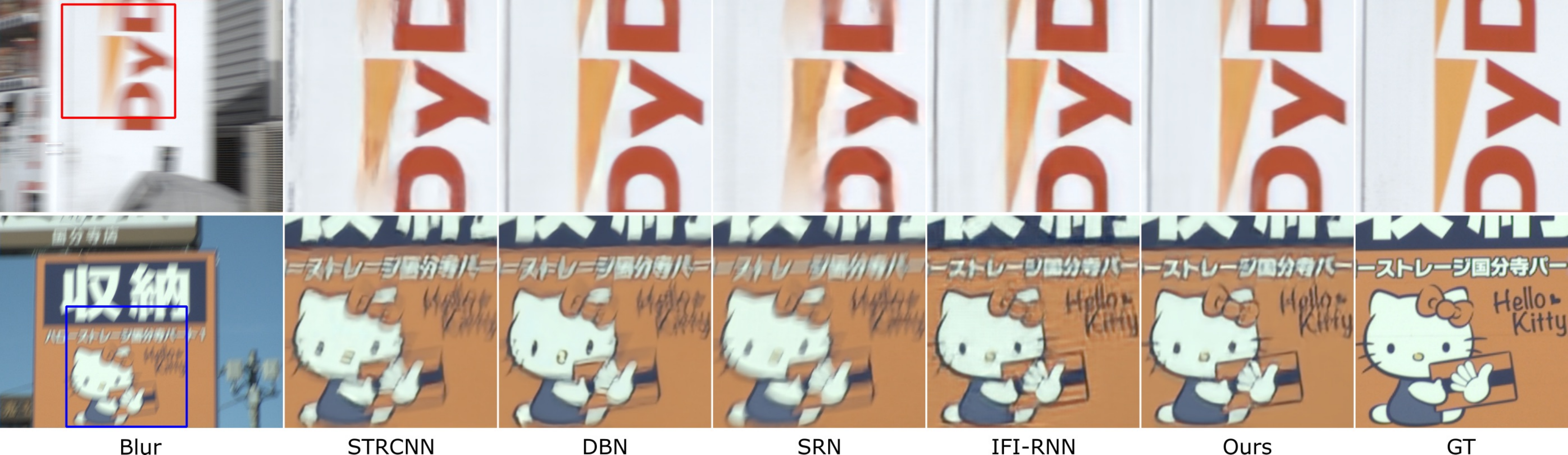}}
	\hfil
	\subfloat[Visual results on BSD (3ms--24ms)]{\includegraphics[width=\textwidth]{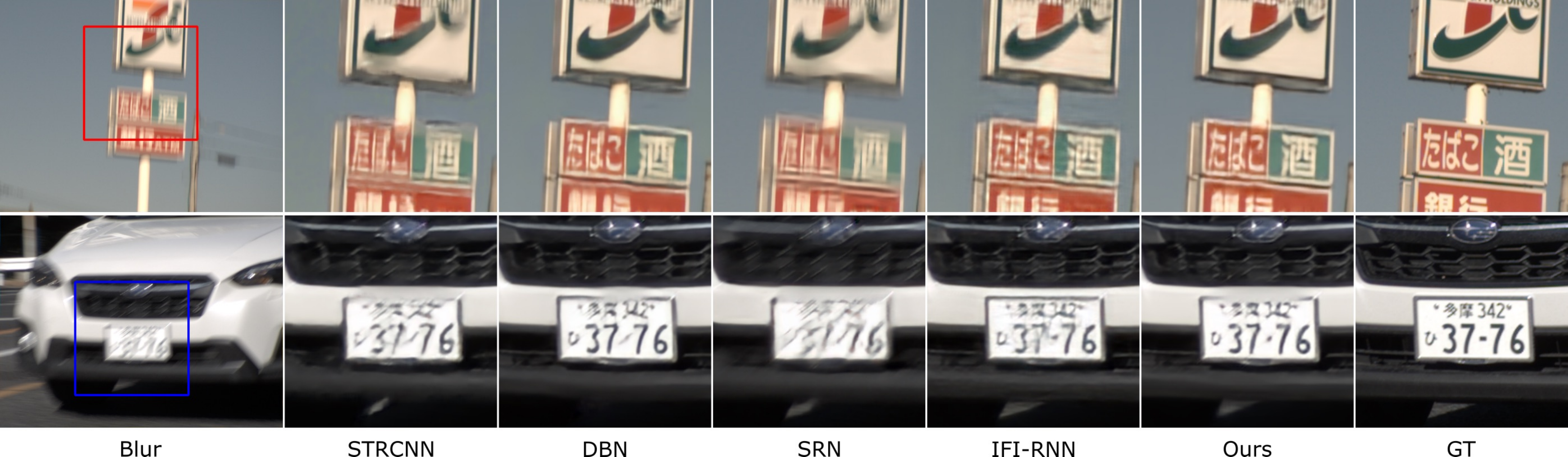}}
	\hfil
	\caption{Visual comparisons on different settings of the proposed BSD.}
	\label{fig:bsd_results}
\end{figure*}

We also conduct comparison experiments on the proposed BSD dataset with different blur intensity settings. The visual results for 1ms--8ms, 2ms-16ms, and 3ms--24ms are illustrated in Fig.~\ref{fig:bsd_results}(a), Fig.~\ref{fig:bsd_results}(b) and Fig.~\ref{fig:bsd_results}(c), respectively. The proposed method ($B15C80$) achieves more visually appealing results in all settings of the real-world deblurring dataset. The quantitative results are shown as Table~\ref{table:bsd}. It indicates that the dataset setting with longer exposure time, \textit{i.e.}, higher blur intensity, is more difficult to restore. Our method achieves the best PSNR and SSIM scores of 33.36dB and 0.937 for 1ms--8ms setting, while the best scores of 31.39dB and 0.296 for 3ms--24ms setting. Both qualitative and quantitative results verify the effectiveness of our method on real-world video deblurring task. Also, comparison results with huge models with pretrained optical flow estimator are presented in the supplemental materials.

\subsection{Dataset Cross-validation}

\subsubsection{Qualitative cross-validation with synthetic datasets}

\begin{figure*}[!ht]
	\centering
	\subfloat[Cross-validation from BSD to GOPRO]{\includegraphics[height=0.4\textwidth]{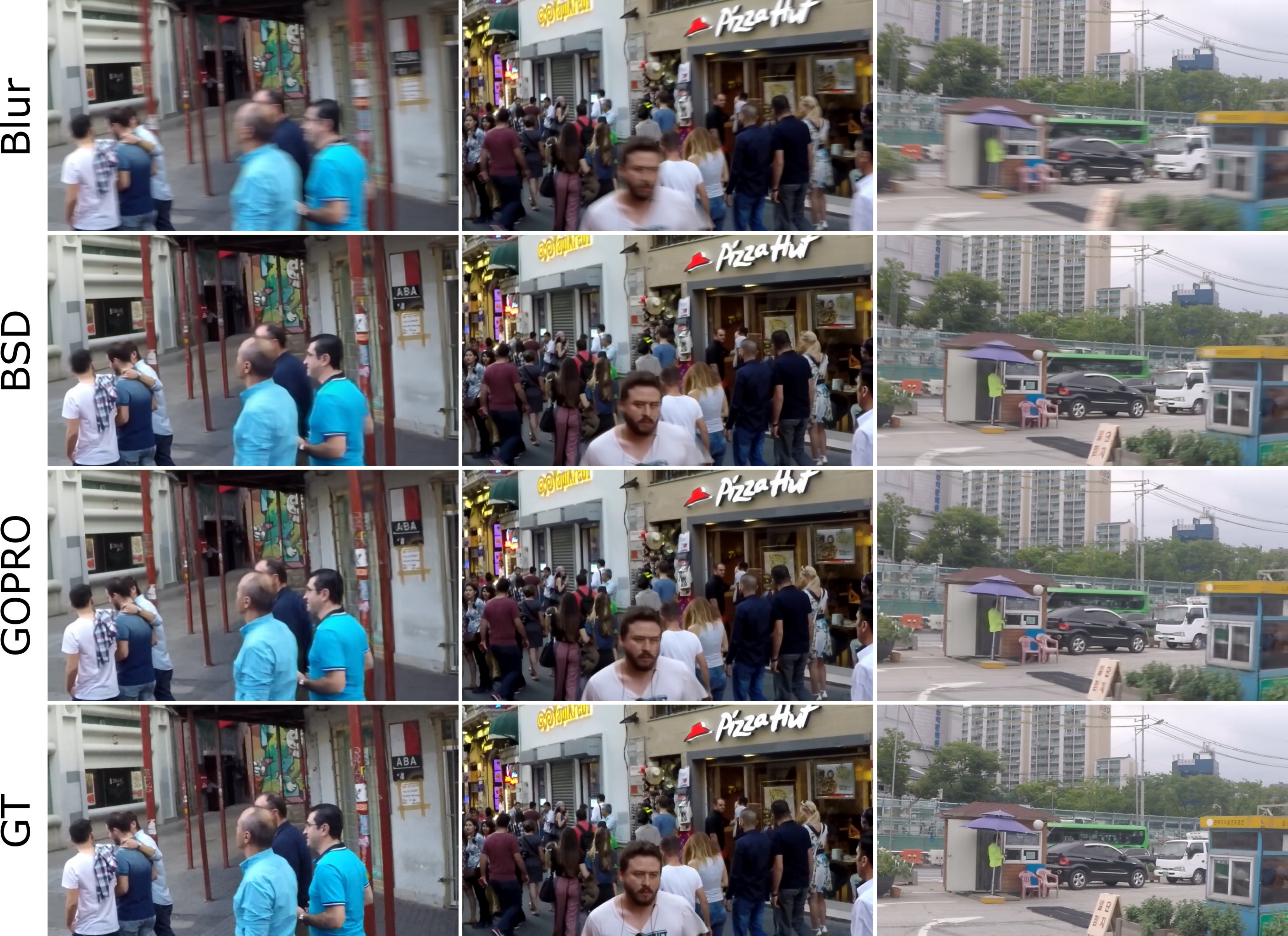}
		\label{fig:bsd2gopro}}
	\hfil
	\subfloat[Cross-validation from GOPRO to BSD]{\includegraphics[height=.4\textwidth]{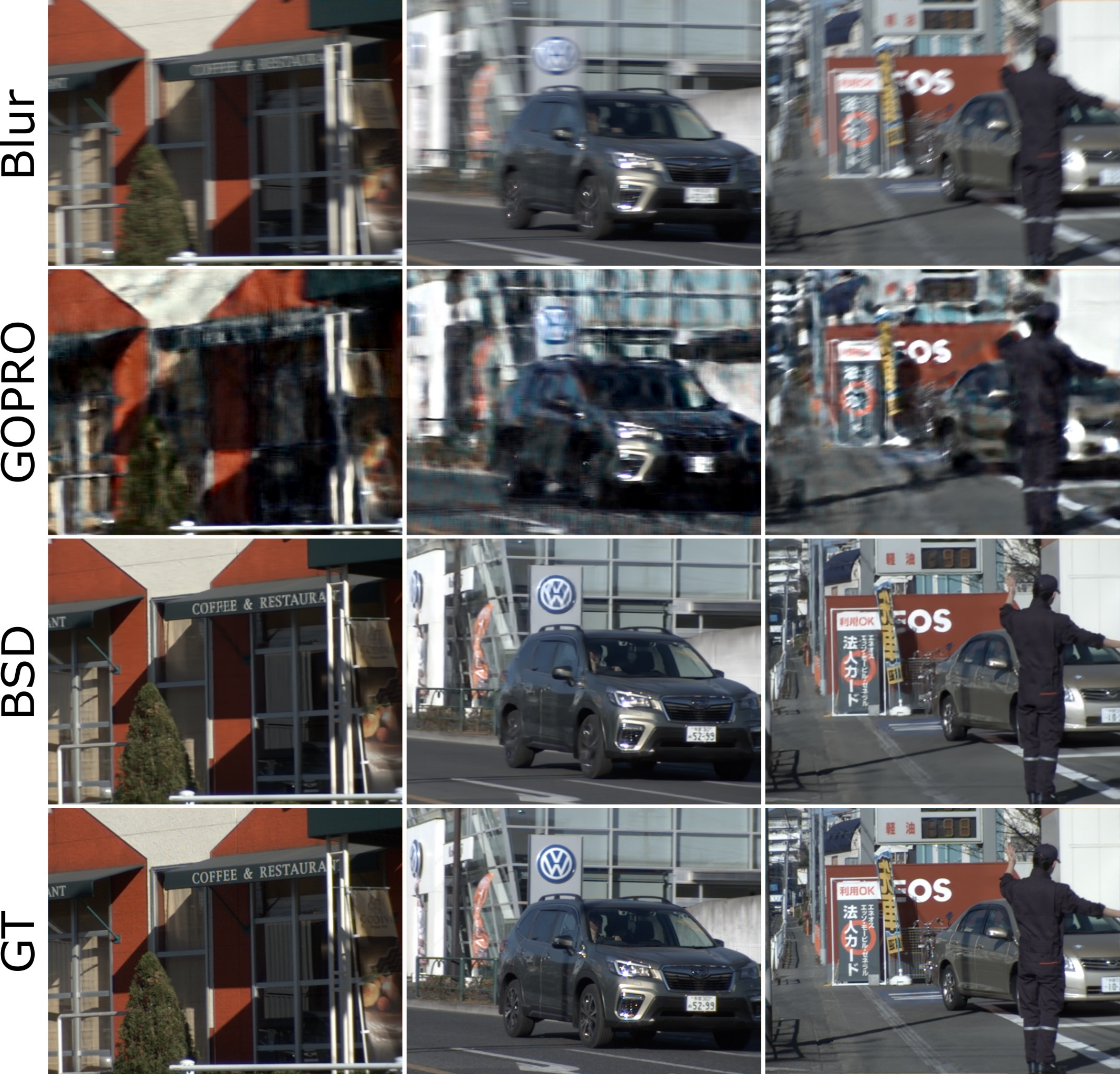}
		\label{fig:gopro2bsd}}
	\caption{Cross-validation between BSD and GOPRO.}
	\label{fig:cross_bsd_gopro}
\end{figure*}

\begin{figure*}[ht]
	\centering
	\subfloat[Cross-validation from BSD to REDS]{\includegraphics[height=.4\textwidth]{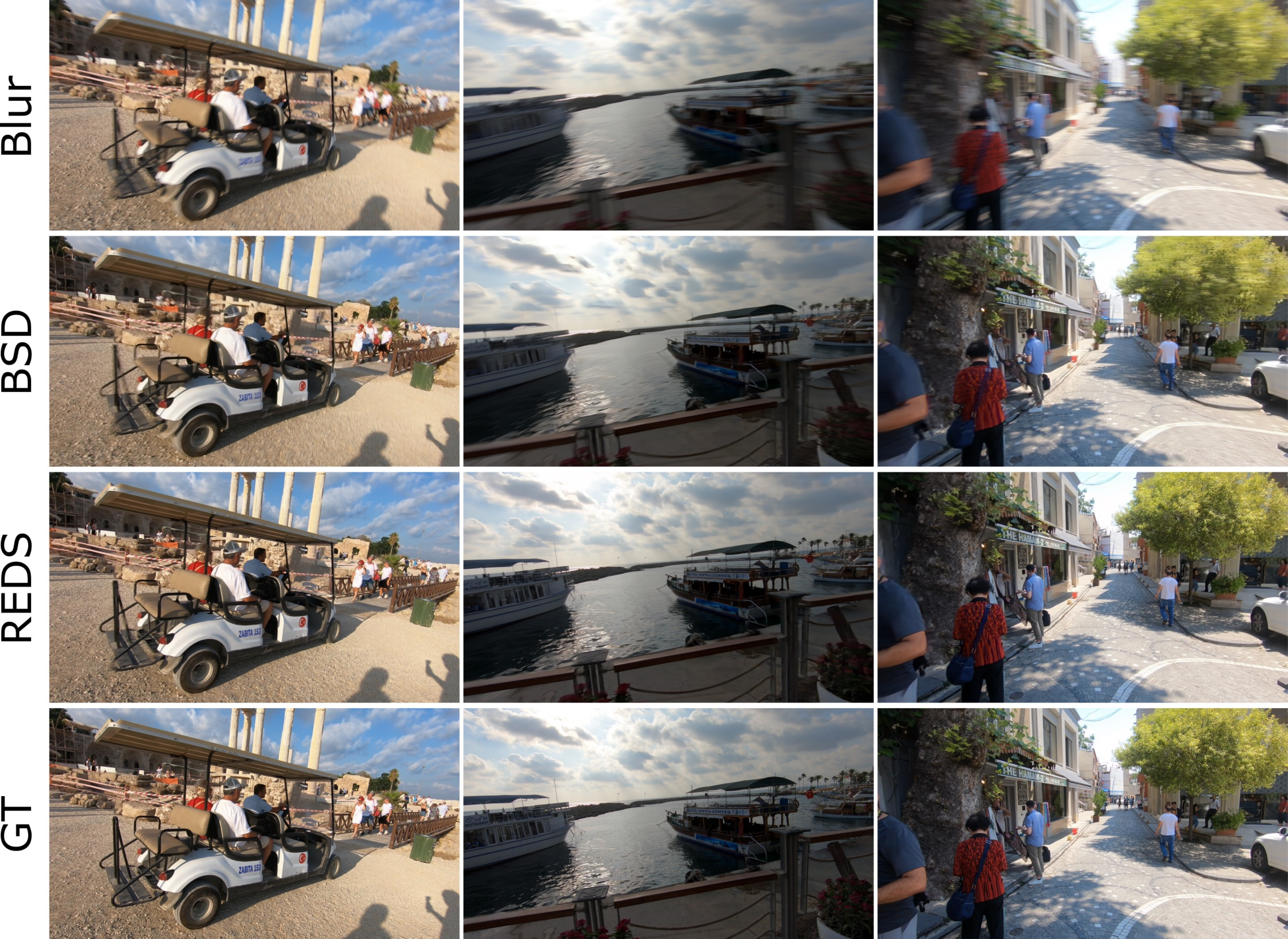}
		\label{fig:bsd2reds}}
	\hfil
	\subfloat[Cross-validation from REDS to BSD]{\includegraphics[height=.4\textwidth]{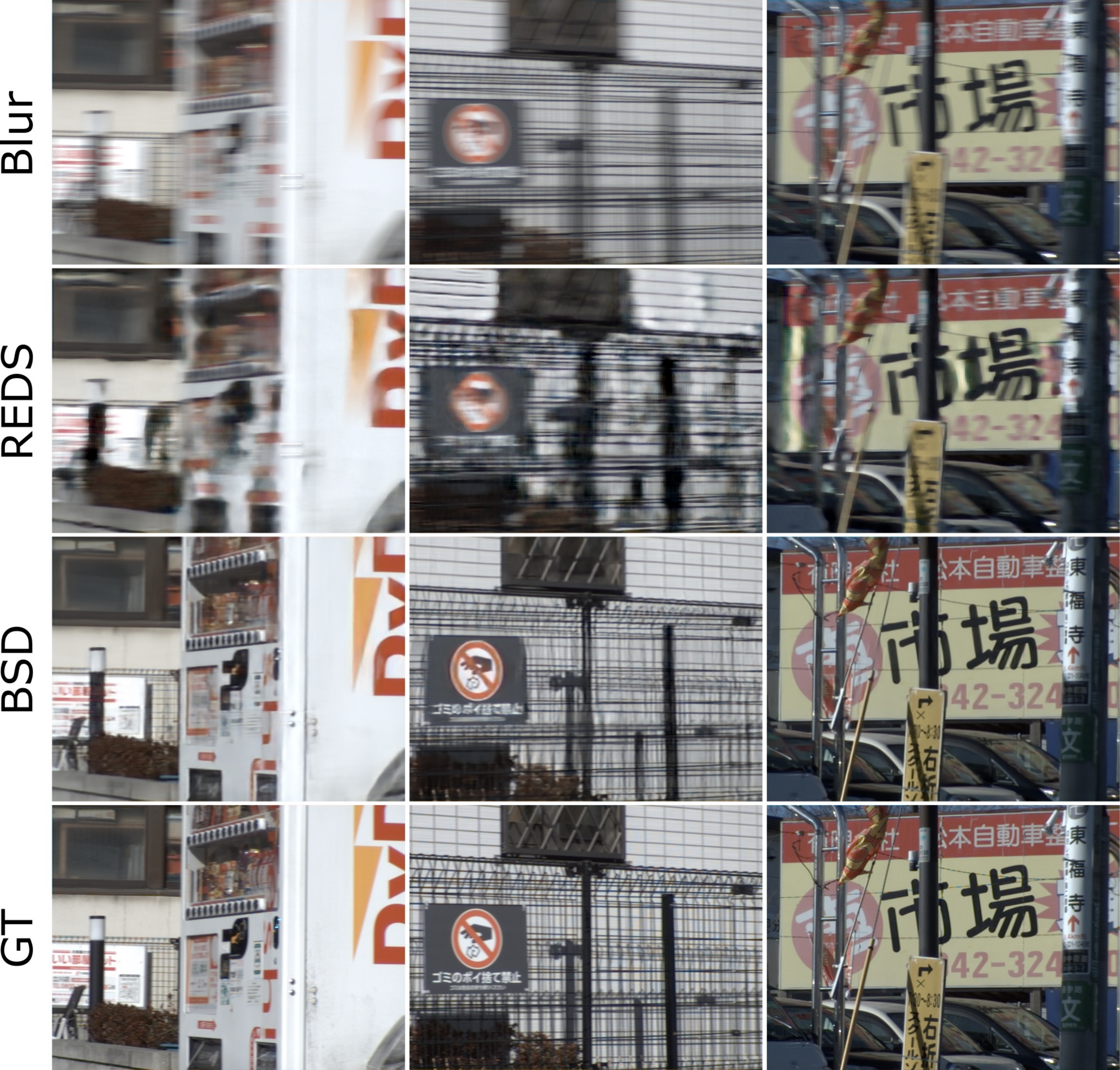}
		\label{fig:reds2bsd}}
	\caption{Cross-validation between BSD and REDS.}
	\label{fig:cross_bsd_reds}
\end{figure*}

\begin{figure*}[ht]
	\centering
	\subfloat[Cross-validation from BSD to DVD]{\includegraphics[height=.4\textwidth]{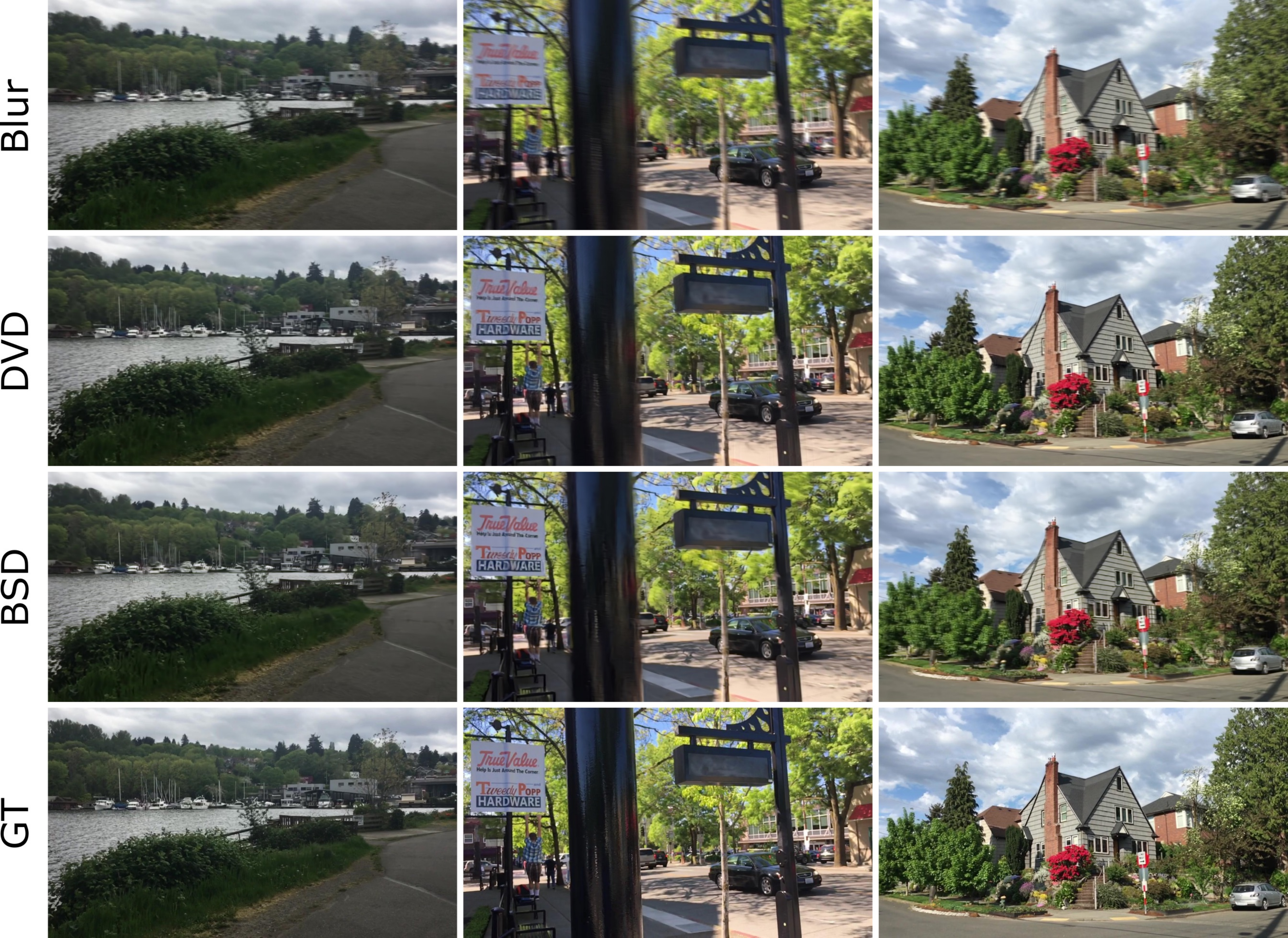}
		\label{fig:bsd2reds}}
	\hfil
	\subfloat[Cross-validation from DVD to BSD]{\includegraphics[height=.4\textwidth]{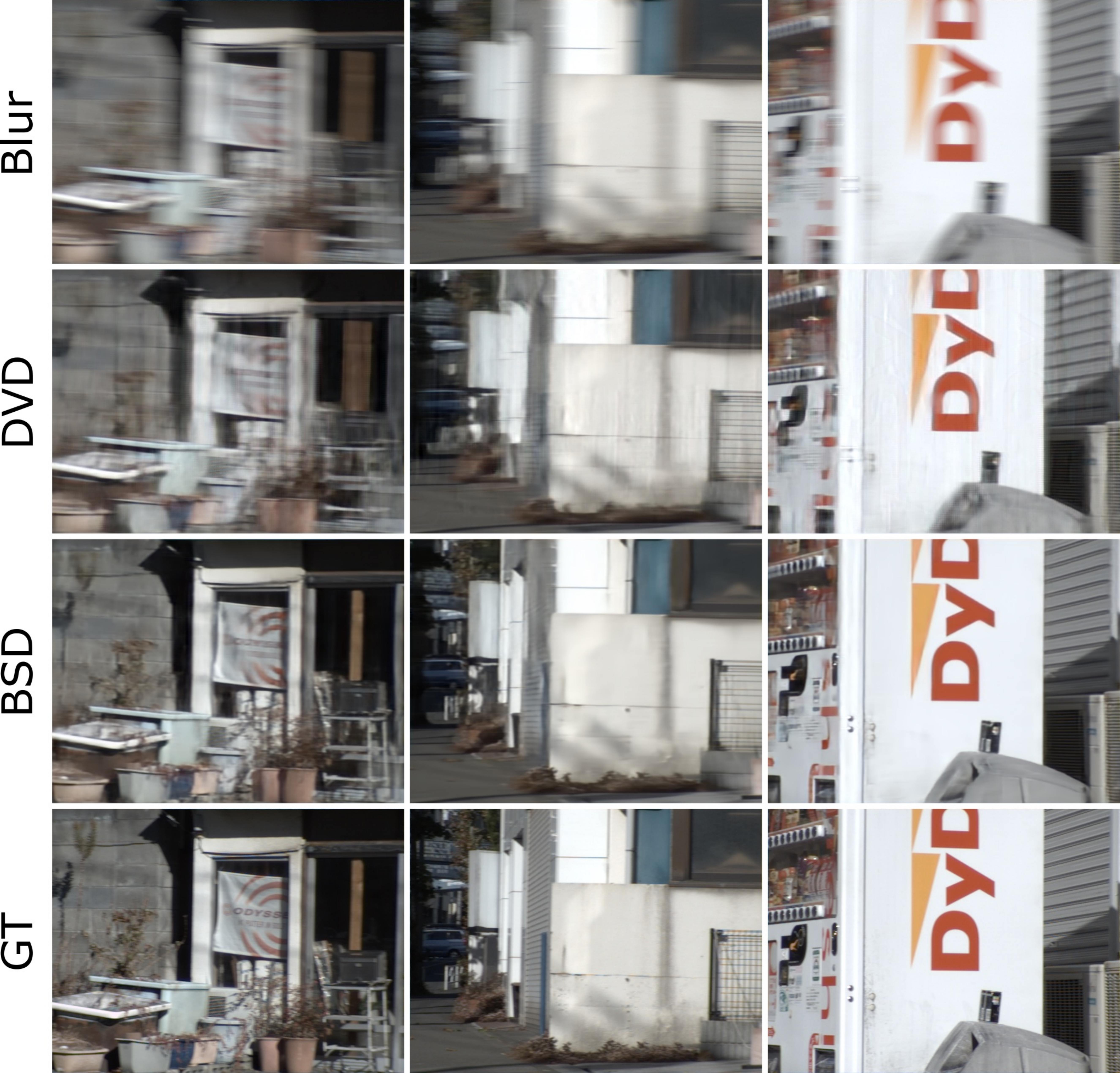}
		\label{fig:reds2bsd}}
	\caption{Cross-validation between BSD and DVD.}
	\label{fig:cross_bsd_dvd}
\end{figure*}

To verify the advantages of using real-world data for training, we further conduct cross-validation between BSD and GOPRO. The predicted results for GOPRO by using our model ($B15C80$) trained on BSD (2ms--16ms) are illustrated in Fig.~\ref{fig:cross_bsd_gopro}(a); while the predicted results for BSD (2ms--16ms) by using our model trained on GOPRO are illustrated in Fig.~\ref{fig:cross_bsd_gopro}(b). The comparison in Fig.~\ref{fig:cross_bsd_gopro} demonstrates that the model trained on real-world dataset has much better generalization ability than the model trained on synthetic dataset. The model trained on BSD has decent deblurring performance on GOPRO. In contrast, the model trained on GOPRO cannot deblur well on BSD but introduce severe artifacts. The above observations also apply to the other two BSD settings, 1ms--8ms and 3ms--24ms. We also show the results of cross-validation between BSD and REDS~\cite{nah2019ntire} in Fig.~\ref{fig:cross_bsd_reds}, as well as BSD and DVD~\cite{su2017deep} in Fig.~\ref{fig:cross_bsd_dvd}. The results demonstrate that REDS dataset with higher quality images also suffers from the same issue. While the model trained with DVD performed better than the model trained with the GOPRO dataset or the REDS dataset when tested on real-world data. However, even if DVDs perform better than other synthetic datasets, the trained model still produces undesired artifacts when processing real-world images with certain strongly blurred regions. This validates that real datasets are still the preferred choice for training deblurring models.

\begin{table*}[ht]
\caption{Dataset cross-validation in terms of PSNR/SSIM. We choose the 2ms16ms setting for the BSD dataset. The setting of ESTRNN is (B15C80).}
  \label{tab:cross}
  \setlength{\tabcolsep}{14pt}
  \centering
  \small
  \begin{tabular}{lcccc}
    \toprule
     & BSD & GOPRO & REDS & DVD\\
    \midrule
    Blur & 26.64/0.818 & 25.47/0.785 & 26.28/0.769 & 27.23/0.813\\
    \midrule
    BSD & 31.95/0.925 & 26.46/0.817 & 27.00/0.801 & 27.88/0.844 \\
    GOPRO & 19.48/0.598 & 31.27/0.903 & 28.21/0.829 & 28.90/0.869\\
    REDS & 24.14/0.773 & 28.43/0.869 & 32.82/0.915 & 28.06/0.848\\
    DVD & 28.67/0.875 & 26.57/0.820 & 26.64/0.787 & 30.68/0.897\\
    \bottomrule
  \end{tabular}
\end{table*}

\subsubsection{Quantitative cross-validation with synthetic datasets}
We also implement the quantitatively comparison for dataset cross-validation, as illustrated in Table.~\ref{tab:cross}. We show the bottom-score for each dataset in the first row. This bottom-score is calculated by using the original blur input without any processing and the corresponding sharp ground-truth. If the scores calculated from the processed images of the model are lower than the bottom-score, it can be considered that the model has a negative impact on the test data set in general. We can find that the model trained using the BSD can consistently obtain at least positive gains on the synthetic datasets. However, the model trained using the GOPRO and REDS will result in a lower performance than the bottom-score of the BSD dataset. The model trained on DVD can also achieve positive gain on BSD, which is consistent with the observation of qualitative results.

\begin{figure*}[!ht]
	\centering
	\subfloat[Cross-validation from BSD to high-fps synthetic dataset]{\includegraphics[height=0.4\textwidth]{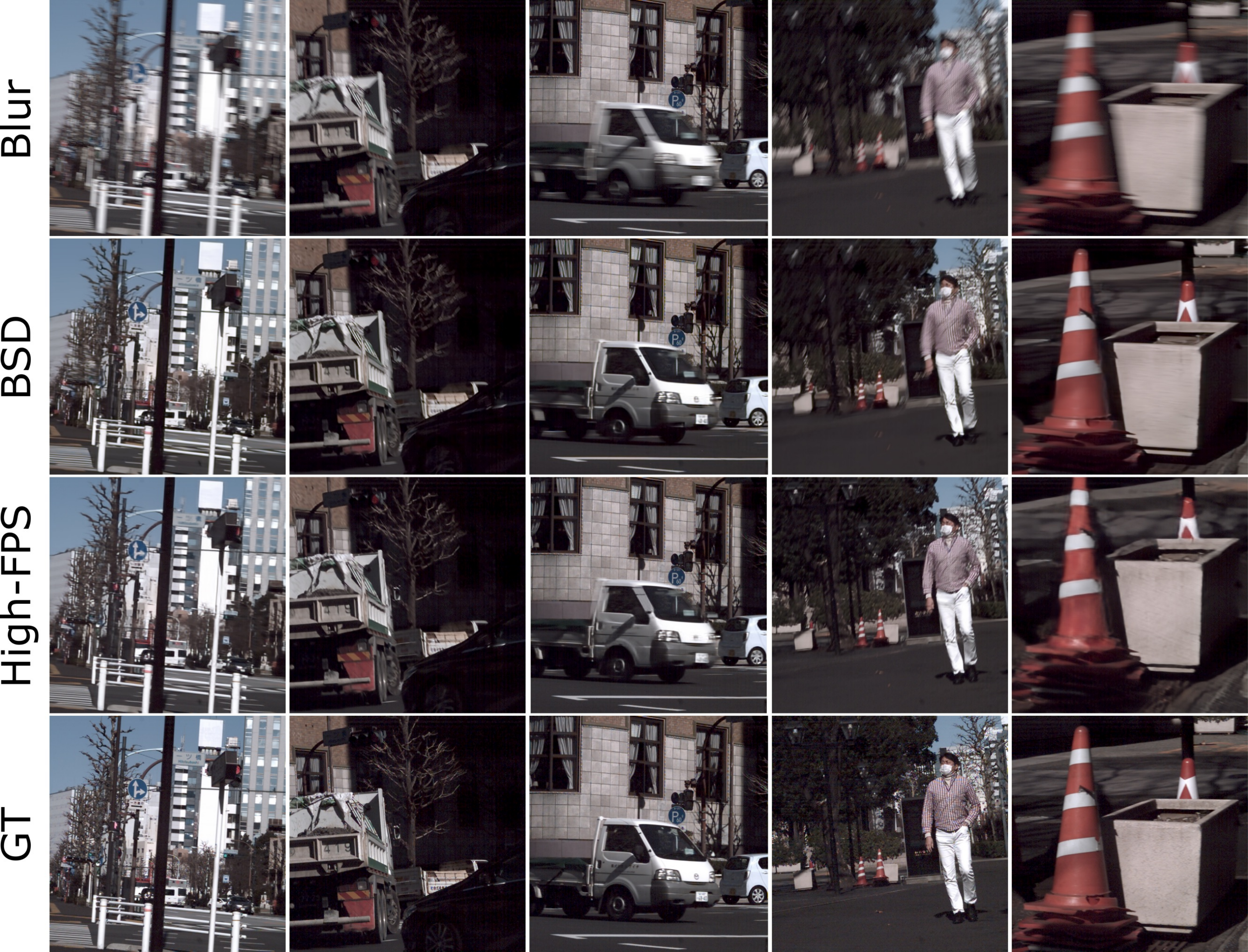}
		\label{fig:bsd2highfps}}
	\hfil
	\subfloat[Cross-validation from high-fps synthetic dataset to BSD]{\includegraphics[height=.4\textwidth]{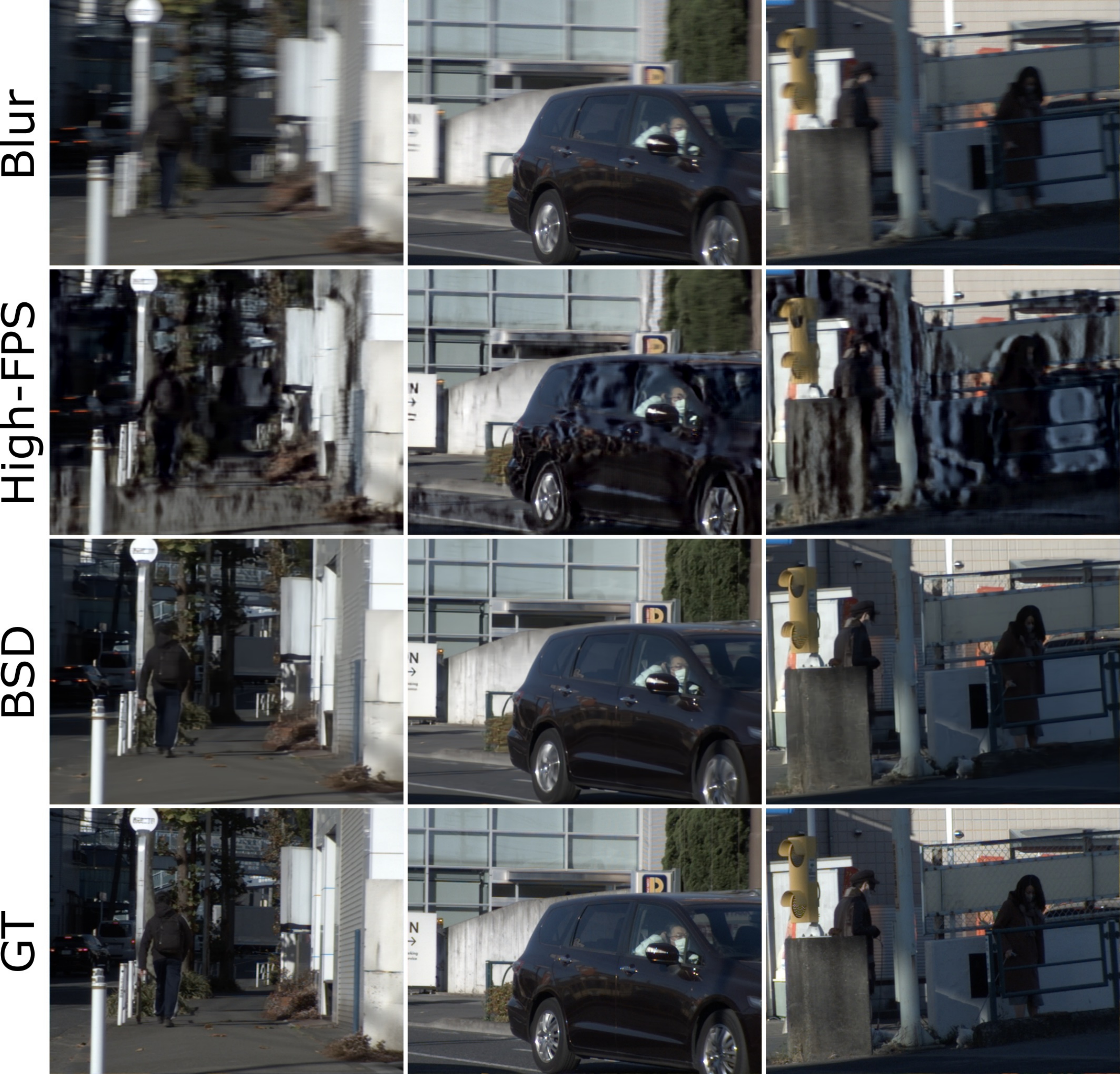}
		\label{fig:highfps2bsdd}}
	\caption{Cross-validation between BSD and high-fps synthetic dataset.}
	\label{fig:cross_bsd_highfps}
\end{figure*}

\begin{figure*}[h]
	\centering
	\includegraphics[width=\textwidth]{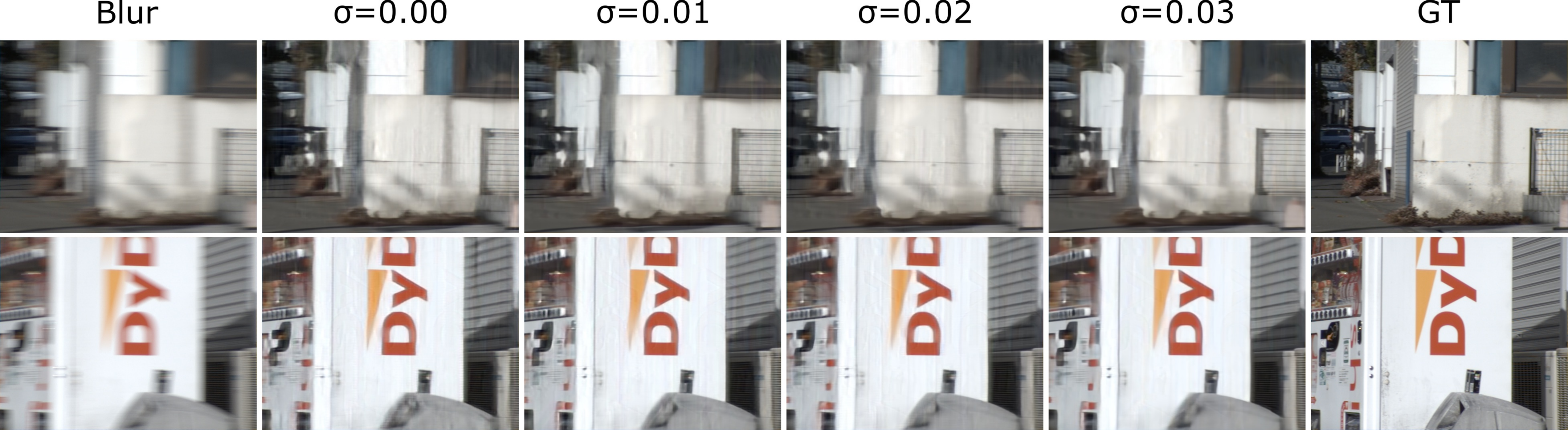}
	\caption{BSD test results of models (ESTRNN B15C80) trained by using DVD with additional Gaussian noise.}
	\label{fig:dvd2bsd_noise}
\end{figure*}

\subsubsection{Possible factors affecting the generalization ability of synthetic data}
The key difference between synthetic data pairs and real-world data pairs is that synthetic data is averaging with discrete images to simulate the formation of blur. We believe that the continuity of the discrete images used for synthesis and the noise distribution of the synthetic blur are two important factors that affect the generalization ability of the synthetic dataset.

First, we assume one of the reasons for causing the unsatisfactory results of the model trained on synthetic dataset is the insufficient frame rate of the original captured high-fps video (\SI{240}{fps} for GOPRO and \SI{120}{fps} for REDS). Although video frame interpolation algorithms can be used to increase the frame rate, it is questionable whether the interpolated frames match the distribution of natural images.

To figure out how insufficient frame rate affects the quality of the synthetic deblurring dataset, we use a camera with a much higher frame rate (\SI{2000}{fps}) to record videos for making a new synthetic dataset. In this case, the readout time is negligible and the exposure time is very close to \SI{0.5}{\milli\second}, thus video interpolation is not needed to supplement the missing information. We build a high-fps synthetic dataset with the same setting as BSD (2ms--16ms, 15fps) using the captured \SI{2000}{fps} videos. Then, we conduct another cross-validation experiment between BSD (2ms--16ms) and the synthetic high-fps dataset to investigate whether a high enough frame rate could solve the problem of poor migration of the synthetic dataset. The results are illustrated in Fig.~\ref{fig:cross_bsd_highfps}. Basically, the experimental results are consistent with the cross-validation between BSD and GOPRO, i.e., the model trained on BSD can work well on the synthetic high-fps dataset (see Fig.~\ref{fig:cross_bsd_highfps}(a)), but not vice-versa (see Fig.~\ref{fig:cross_bsd_highfps}(b)). Therefore, simply satisfying sufficient continuity cannot improve the generalization ability of the synthetic data.

From the perspective of noise distribution of blurred images, the present synthesis method cannot simulate the noise generated in real blurred images at the RAW acquisition and ISP stages. The process of averaging successive sharp images suppresses some of the noise. Thus, a likely reason for the one-sided relationship is that the model trained on BSD has seen more complex noise models so it can handle synthetic data with noise suppressed, which is the simpler case, while the reverse is difficult.

In addition, we trained a series of models on DVD by adding Gaussian noise to the RGB space of blurred images with different standard deviations $\sigma$. The visual results in Fig.~\ref{fig:dvd2bsd_noise} indicate that adding the appropriate noise can slightly improve artifact, but it is still far from eradicating it. Starting from RAW space to synthesize blur and considering the appropriate noise model should be a promising direction.

The above experiments demonstrate the significance of the real-world dataset and indicate that it is not trivial to produce a high-quality deblurring datasets by synthetic methods.

\begin{figure*}[ht]
	\centering
    \includegraphics[width=\linewidth]{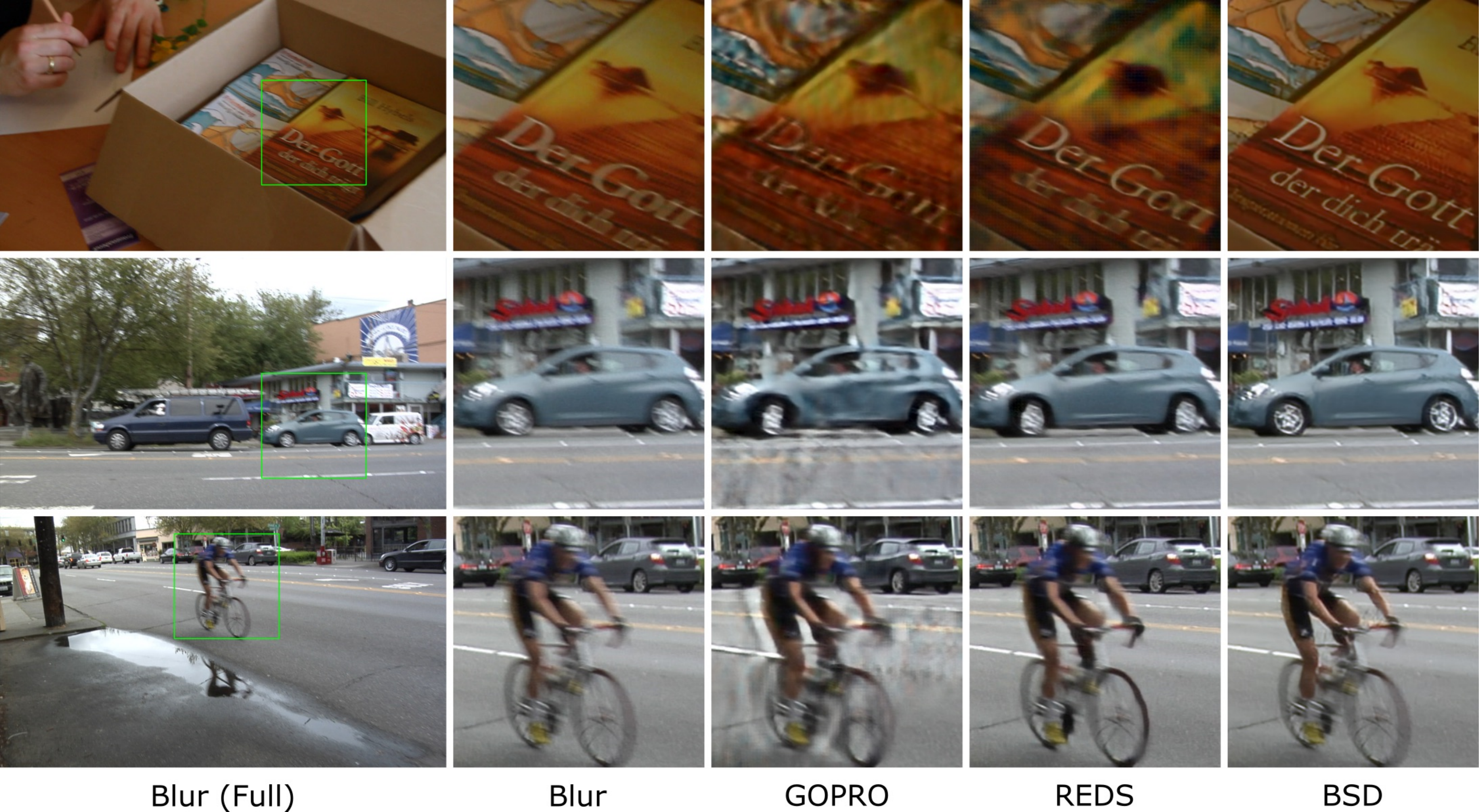}
	\caption{Qualitative results on real-world videos from~\cite{cho2012video,su2017deep}}
	\label{fig:dvd}
\end{figure*}

\subsection{Test on Third-part Real-world Videos}
Our real-world dataset BSD is obtained under some specific exposure settings. To verify the generalization ability of BSD, we further verify our model trained by using BSD on the commonly used real-world videos from DVD~\cite{cho2012video,su2017deep}, as illustrated in Fig.~\ref{fig:dvd}. We can see that our model trained on the synthetic datasets, GOPRO~\cite{nah2017deep} and REDS~\cite{nah2019ntire} inevitably introduce undesired artifacts to the final results, such as the cover of the book, the wheel of the car, and the body of the man. Our model trained on BSD (2ms16ms) works well in most of these cases. It is worth noting that in the bicycle example, even though the model trained on BSD cannot recover the very challenging blurred bicycle wheel, the model does not force to add some artifacts. We believe that this is advantageous in terms of visual perception and can be further addressed by expanding the dataset.

\begin{figure*}[ht]
	\centering
	\includegraphics[width=\textwidth]{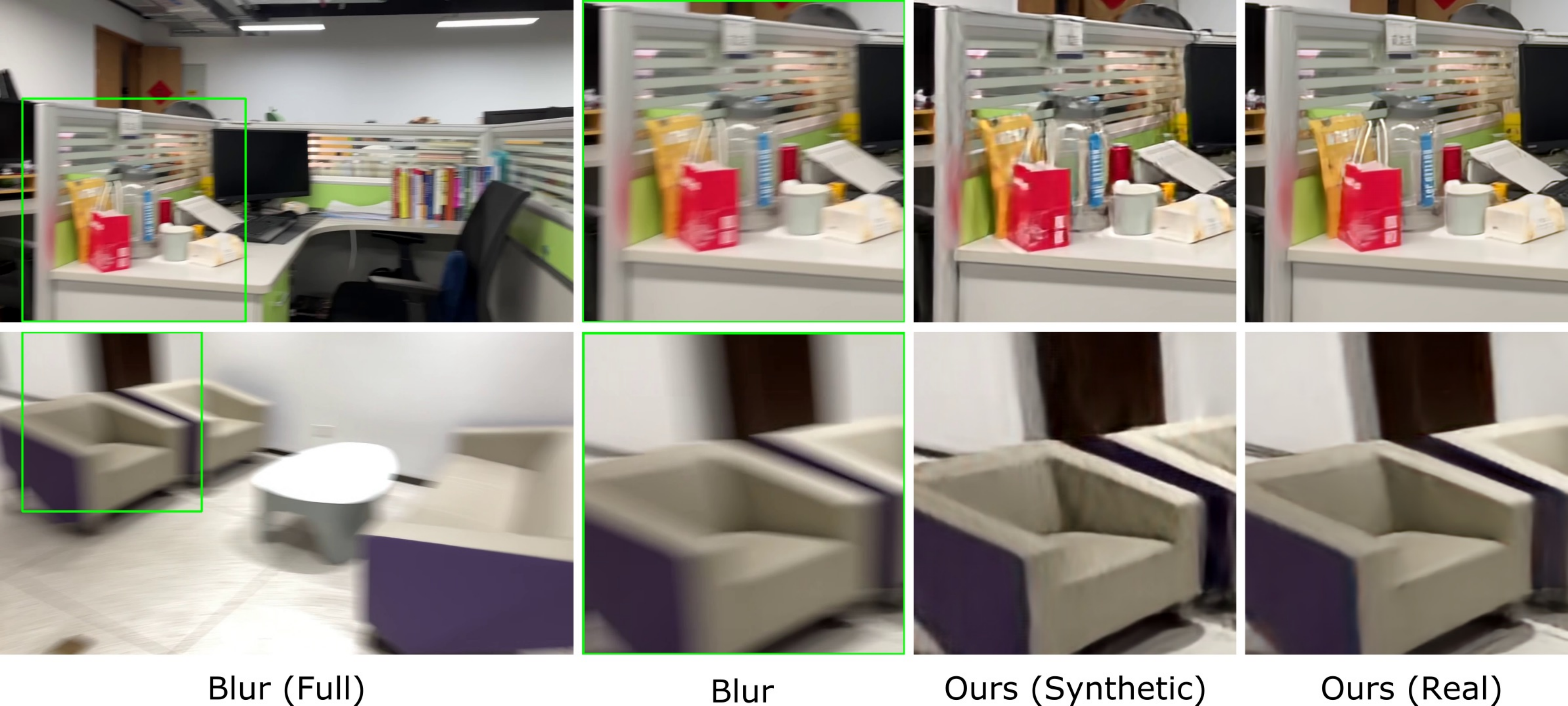}
	\caption{Testing on real-world blurry videos from iPhone 13. Ours (Synthetic) denotes the results of ESTRNN trained on synthetic dataset GOPRO. Ours (real) denotes the results of ESTRNN trained on real-world dataset BSD (2ms-16ms).}
	\label{fig:cross_device}
\end{figure*}

We also test our model by shooting blurred videos without any specific constraints using iPhone 13. The predicted results from our model trained on synthetic dataset GOPRO and on our real-world dataset BSD are illustrated in Fig.~\ref{fig:cross_device}. It is clear that the model trained on real-world data successfully restored the latent images and generated clearer details such as the handle of the bag. While the model trained on synthetic dataset suffered from unsharp details and undesired artifacts.

\section{Conclusion}
In this paper, we proposed a novel RNN-based method for more computationally efficient video deblurring. Residual dense blocks were adopted to the RNN cell to generate hierarchical features from current frame for better restoration. Moreover, to make full use of the spatio-temporal correlation, our model utilized the global spatio-temporal fusion module for fusing the effective components of hierarchical features from past and future frames. Furthermore, we have developed a beam-splitter acquisition system and contributed the first real-world dataset for image/video deblurring tasks. The experimental results show that our model is more computationally efficient for video deblurring, which can achieve better performance with less computational cost. Cross-validation experiments between real-world and synthetic datasets demonstrate the high generality of the proposed BSD dataset.

\begin{acknowledgements}
This work was supported in part by JSPS KAKENHI Grant Numbers JP20H05951 and JP20H05953.
\end{acknowledgements}

%
%

\bibliographystyle{spmpsci}      
\bibliography{main}   


\end{document}